\documentclass{article}
\usepackage[final]{neurips_2019}
\PassOptionsToPackage{sort&compress}{natbib}
\setcitestyle{numbers,square,comma,sort}
\usepackage{microtype}
\usepackage{graphicx}
\usepackage{subfigure}
\usepackage{booktabs} 
\usepackage{wrapfig}

\usepackage{hyperref}

\usepackage{url}
\usepackage[font=small]{caption}
\usepackage{wrapfig}
\usepackage{textcomp}
\usepackage{blindtext}

\usepackage{multirow}
\usepackage{dsfont}
\usepackage{multicol}
\usepackage{amsmath}
\usepackage{amsthm}
\usepackage{amssymb}
\usepackage{bm}
\usepackage{mathrsfs}
\usepackage{xcolor}
\usepackage{colortbl}
\usepackage{enumitem}
\usepackage{mathtools}

\usepackage{color}
\usepackage{soul}

\usepackage{setspace}
\AtBeginDocument{%
  \addtolength\abovedisplayskip{-0.2\baselineskip}%
  \addtolength\belowdisplayskip{-0.2\baselineskip}%
} 

\usepackage{pgfplots}
\pgfplotsset{
	compat=newest,
	plot coordinates/math parser=false,
	tick label style={font=\scriptsize, /pgf/number format/fixed},
	label style={font=\small},
	every axis/.append style={
		tick align=outside,
		clip mode=individual,
		scaled ticks=false,
		thick,
		tick style={semithick, black}
	}
}

\makeatletter
\renewenvironment{proof}[1][\proofname]{\par
  \vspace{-\topsep}
  \pushQED{\qed}%
  \normalfont
  \topsep0pt \partopsep0pt 
  \trivlist
  \item[\hskip\labelsep
        \itshape
    #1\@addpunct{.}]\ignorespaces
}{%
  \popQED\endtrivlist\@endpefalse
  \addvspace{6pt plus 6pt} 
}
\makeatother

\usepackage[toc,titletoc,page]{appendix}
\let\appendixpagenameorig\appendixpagename
\renewcommand{\appendixpagename}{\Large\appendixpagenameorig}

\pgfkeys{/pgf/number format/.cd, set thousands separator={\,}}

\usetikzlibrary{calc}
\usepgfplotslibrary{external}
\newlength\figurewidth
\newlength\figureheight

\newlength\sbsfigurewidth
\newlength\sbsfigureheight

\usepackage{xspace}

\newcommand{\given}{\mid}

\newcommand{\mat}[1]{\bm{\mathrm{#1}}}
\renewcommand{\vec}[1]{\bm{\mathrm{#1}}}
\newcommand{\acro}[1]{\textsc{\MakeLowercase{#1}}}

\newtheorem{definition}{Definition}
\newtheorem{theorem}{Theorem}

\newcommand{\gp}{\acro{GP}\xspace}
\newcommand{\bo}{\acro{BO}\xspace}

\newcommand{\DAG}{\acro{DAG}\xspace}
\newcommand{\DAGs}{\acro{DAG}s\xspace}
\newcommand{\VAE}{\acro{VAE}\xspace}
\newcommand{\DVAE}{\acro{D-VAE}\xspace}
\newcommand{\GVAE}{\acro{G-VAE}\xspace}
\newcommand{\GNN}{\acro{GNN}\xspace}
\newcommand{\GNNs}{\acro{GNN}s\xspace}
\newcommand{\CNN}{\acro{CNN}\xspace}
\newcommand{\CNNs}{\acro{CNN}s\xspace}
\newcommand{\RNN}{\acro{RNN}\xspace}
\newcommand{\RNNs}{\acro{RNN}s\xspace}
\newcommand{\NAS}{\acro{NAS}\xspace}
\newcommand{\ENAS}{\acro{ENAS}\xspace}
\newcommand{\BNSL}{\acro{BNSL}\xspace}
\newcommand{\BIC}{\acro{BIC}\xspace}
\newcommand{\MLPs}{\acro{MLP}s\xspace}
\newcommand{\MLP}{\acro{MLP}\xspace}
\newcommand{\GRU}{\acro{GRU}\xspace}
\newcommand{\GRUs}{\acro{GRU}s\xspace}
\newcommand{\GraphRNN}{\acro{G}raph\acro{RNN}\xspace}
\newcommand{\SVAE}{\acro{S-VAE}\xspace}
\newcommand{\GCN}{\acro{GCN}\xspace}
\newcommand{\dgmg}{\acro{D}eep\acro{GMG}\xspace}
\newcommand{\RMSE}{\acro{RMSE}\xspace}
\newcommand{\SGP}{\acro{SGP}\xspace}
\allowdisplaybreaks

\usepackage[utf8]{inputenc} 
\usepackage[T1]{fontenc}    
\usepackage{hyperref}       
\usepackage{url}            
\usepackage{booktabs}       
\usepackage{amsfonts}       
\usepackage{nicefrac}       
\usepackage{microtype}      

\usepackage{setspace}
\AtBeginDocument{%
  \addtolength\abovedisplayskip{-0.2\baselineskip}%
  \addtolength\belowdisplayskip{-0.2\baselineskip}%
} 

\usepackage{titlesec}
\titlespacing\subsection{0pt}{4pt plus 2pt minus 2pt}{3pt plus 2pt minus 2pt}

\title{D-VAE: A Variational Autoencoder for Directed Acyclic Graphs}

\author{%
  Muhan Zhang, Shali Jiang, Zhicheng Cui, Roman Garnett, Yixin Chen\\
  Department of Computer Science and Engineering\\
  Washington University in St. Louis\\
  \texttt{\{muhan, jiang.s, z.cui, garnett\}@wustl.edu}, \texttt{chen@cse.wustl.edu} \\
}

\begin{document}

\maketitle
\begin{abstract}
Graph structured data are abundant in the real world. Among different graph types, directed acyclic graphs ({\DAGs}) are of particular interest to machine learning researchers, as many machine learning models are realized as computations on \DAGs, including neural networks and Bayesian networks. In this paper, we study deep generative models for \DAGs, and propose a novel \DAG variational autoencoder (\DVAE). To encode \DAGs into the latent space, we leverage graph neural networks. We propose an asynchronous message passing scheme that allows encoding the computations on \DAGs, rather than using existing simultaneous message passing schemes to encode local graph structures. We demonstrate the effectiveness of our proposed \DVAE through two tasks: neural architecture search and Bayesian network structure learning. Experiments show that our model not only generates novel and valid \DAGs, but also produces a smooth latent space that facilitates searching for \DAGs with better performance through Bayesian optimization.
\end{abstract}

	
\section{Introduction}
\label{intro}
Many real-world problems can be posed as optimizing of a directed acyclic graph (\DAG) representing some computational task. For example, the architecture of a neural network is a \DAG. The problem of searching optimal neural architectures is essentially a \DAG optimization task. Similarly, one critical problem in learning graphical models -- optimizing the connection structures of Bayesian networks \cite{koller2009probabilistic}, is also a \DAG optimization task. \DAG optimization is pervasive in other fields as well. In electronic circuit design, engineers need to optimize \DAG circuit blocks not only to realize target functions, but also to meet specifications such as power usage and operating temperature.



\DAG optimization is a hard problem. Firstly, the evaluation of a \DAG's performance is often time-consuming (e.g., training a neural network).
Secondly, state-of-the-art black-box optimization techniques such as simulated annealing and Bayesian optimization primarily operate in a continuous space, thus are not directly applicable to \DAG optimization due to the discrete nature of \DAGs. In particular, to make Bayesian optimization work for discrete structures, we need a kernel to measure the similarity between discrete structures as well as a method to explore the design space and extrapolate to new points. Principled solutions to these problems are still lacking.

Is there a way to circumvent the trouble from discreteness? The answer is yes. If we can \textbf{embed all \DAGs to a continuous space} and make the space relatively smooth, we might be able to directly use principled black-box optimization algorithms to optimize \DAGs in this space, or even use gradient methods if gradients are available. Recently, there has been increased interest in training generative models for discrete data types such as molecules \cite{gomez2018automatic,kusner2017grammar}, arithmetic expressions \cite{kusner2016gans}, source code \cite{gaunt2016terpret}, undirected graphs \cite{li2018learning}, etc. In particular, \citet{kusner2017grammar} developed a grammar variational autoencoder (\GVAE) for molecules, which is able to encode and decode molecules into and from a \textbf{continuous latent space}, allowing one to optimize molecule properties by searching in this well-behaved space instead of a discrete space. Inspired by this work, we propose to also train a variational autoencoder for \DAGs, and optimize \DAG structures in the latent space via Bayesian optimization.

To encode \DAGs, we leverage graph neural networks (\GNNs) \cite{wu2019comprehensive}. Traditionally, a \GNN treats all nodes symmetrically, and extracts local features around nodes by \textbf{simultaneously} passing all nodes' neighbors' messages to themselves. 
However, such a simultaneous message passing scheme is designed to learn local structure features. It might not be suitable for \DAGs, since in a \DAG: 1) nodes are not symmetric, but intrinsically have some ordering based on its dependency structure; and 2) we are more concerned about the computation represented by the entire graph, not the local structures.





In this paper, we propose an \textbf{asynchronous message passing scheme} to encode the computations on \DAGs. The message passing no longer happens at all nodes simultaneously, but respects the computation dependencies (the partial order) among the nodes. For example, suppose node A has two predecessors, B and C, in a \DAG. Our scheme does not perform feature learning for A until the feature learning on B and C are both finished. Then, the aggregated message from B and C is passed to A to trigger A's feature learning. This means, although the message passing is not simultaneous, it is also not completely unordered -- some synchronization is still required. 
We incorporate this feature learning scheme in both our encoder and decoder, and propose \emph{\DAG variational autoencoder} (\DVAE). \DVAE has an excellent theoretical property for modeling \DAGs -- we prove that \DVAE can \textbf{injectively} encode \textbf{computations} on \DAGs. 
This means, we can build a mapping from the discrete space to a continuous latent space so that \textbf{every} \DAG computation has its \textbf{unique} embedding in the latent space, which \textbf{justifies} performing optimization in the latent space instead of the original design space.



Our contributions in this paper are: 
1) We propose \DVAE, a variational autoencoder for \DAGs using a novel asynchronous message passing scheme, which is able to injectively encode computations. 
2) Based on \DVAE, we propose a new \DAG optimization framework which performs Bayesian optimization in a continuous latent space. 
3) We apply \DVAE to two problems, neural architecture search and Bayesian network structure learning. 
Experiments show that \DVAE not only generates novel and valid \DAGs, but also learns smooth latent spaces effective for optimizing \DAG structures.



\section{Related work}\label{main:related}

\noindent \textbf{Variational autoencoder (\VAE)} \cite{kingma2013auto,rezende2014stochastic} provides a framework to learn both a probabilistic generative model $p_\theta(\vec{x}|\vec{z})$ (the decoder) as well as an approximated posterior distribution $q_\phi(\vec{z}|\vec{x})$ (the encoder). \VAE is trained through maximizing the evidence lower bound
\begin{align}
\mathcal{L}(\phi,\theta;\vec{x}) = \mathbb{E}_{\vec{z} \sim q_\phi(\vec{z}|\vec{x})} [\log p_\theta(\vec{x}|\vec{z})] - \mathrm{KL}[q_\phi(\vec{z}|\vec{x}) \| p(\vec{z})].
\label{VAE_loss}
\end{align}
The posterior approximation $q_\phi(\vec{z}|\vec{x})$ 
and the generative model $p_\theta(\vec{x}|\vec{z})$ can in principle take arbitrary parametric forms whose parameters $\phi$ and $\theta$ are output by the encoder and decoder networks. After learning $p_\theta(\vec{x}|\vec{z})$, we can generate new data by decoding latent space vectors $\vec{z}$ sampled from the prior $p(\vec{z})$. For generating discrete data, $p_\theta(\vec{x}|\vec{z})$ is often decomposed into a series of decision steps.

\noindent \textbf{Deep graph generative models} use neural networks to learn distributions over graphs. There are mainly three types:
token-based, adjacency-matrix-based, and graph-based. Token-based models \cite{gomez2018automatic,kusner2017grammar,dai2018syntax} represent a graph as a sequence of tokens (e.g., characters, grammar rules) and model these sequences using \RNNs. 
They are less general since task-specific graph grammars such as \acro{SMILES} for molecules \cite{weininger1988smiles} are required. Adjacency-matrix-based models \cite{simonovsky2018graphvae,you2018graphrnn,de2018molgan,bojchevski2018netgan,ma2018constrained} leverage the proxy adjacency matrix representation of a graph, and generate the matrix in one shot or generate the columns/entries sequentially. 
In contrast, graph-based models \cite{li2018learning,jin2018junction,liu2018constrained,you2018graph} seem more natural, since they operate directly on graph structures (instead of proxy matrix representations) by iteratively adding new nodes/edges to a graph based on the existing graph and node states. In addition, the graph and node states are learned by \textbf{graph neural networks (\GNNs)}, which have already shown their powerful graph representation learning ability on various tasks \cite{duvenaud2015convolutional,li2015gated,kipf2016semi,niepert2016learning,hamilton2017inductive,zhang2018end,zhang2018link,zhang2019inductive}.

\noindent \textbf{Neural architecture search (\NAS)} aims at automating the design of neural network architectures. It has seen major advances in recent years \citep{zoph2016neural,real2017large,elsken2017simple,zoph2018learning,liu2018darts,pham2018efficient}.
See \citet{automl_book} for an overview. 
\NAS methods can be mainly categorized into: 
1) reinforcement learning methods \cite{zoph2016neural,zoph2018learning,pham2018efficient} which train controllers to generate architectures with high rewards in terms of validation accuracy, 
2) Bayesian optimization based methods \cite{kandasamy2018neural} which define kernels to measure architecture similarity and extrapolate the architecture space heuristically, 
3) evolutionary approaches \cite{real2017large,liu2017hierarchical,miikkulainen2019evolving} which use evolutionary algorithms to optimize neural architectures, and 
4) differentiable methods \cite{liu2018darts,cai2018proxylessnas,luo2018neural} which use continuous relaxation/mapping of neural architectures to enable gradient-based optimization. 
In Appendix \ref{related}, we include more detailed discussion on several most related works. 


\noindent \textbf{Bayesian network structure learning (\BNSL)} is to learn the structure of the underlying Bayesian network from observed data \cite{chow1968approximating,gao2017local, pmlr-v80-gao18b, linzner2018cluster}. 
Bayesian network is a probabilistic graphical model encoding conditional dependencies among variables via a \DAG \cite{koller2009probabilistic}.
One main approach for \BNSL is score-based search, i.e., define some ``goodness-of-fit'' score for network structures, and search for one with the optimal score in the discrete design space.
Commonly used scores include \BIC and \acro{BD}eu, mostly based on marginal likelihood \citep{koller2009probabilistic}. 
Due to the \acro{NP}-hardness \cite{chickering1996learning}, however, exact algorithms such as dynamic programming \citep{singh2005finding} or shortest path approaches \citep{yuan2011learning, yuan2013learning} can only solve small-scale problems.
Thus, people have to resort to heuristic methods such as local search and simulated annealing, etc. \citep{chickering1995learning}. \BNSL is still an active research area \cite{gao2017local, linzner2018cluster, silander2018quotient, zheng2018dags, yu2019dag}.

\section{DAG variational autoencoder (D-VAE)}
\label{sec:D-VAE}

In this section, we describe our proposed \DAG variational autoencoder (\DVAE). \DVAE uses an asynchronous message passing scheme to encode and decode \DAGs. 
In contrast to the simultaneous message passing in traditional \GNNs, \DVAE allows encoding \emph{computations} rather than \emph{structures}. 

\begin{definition}\label{computation}
\textbf{(Computation)} Given a set of elementary operations $\mathcal{O}$, a computation $C$ is the composition of a finite number of operations $o \in \mathcal{O}$ applied to an input signal $x$, with the output of each operation being the input to its succeeding operations.
\end{definition}

\begin{wrapfigure}[8]{L}{0.6\textwidth}
\centering
\vspace{-10pt}
\includegraphics[width=0.6\textwidth]{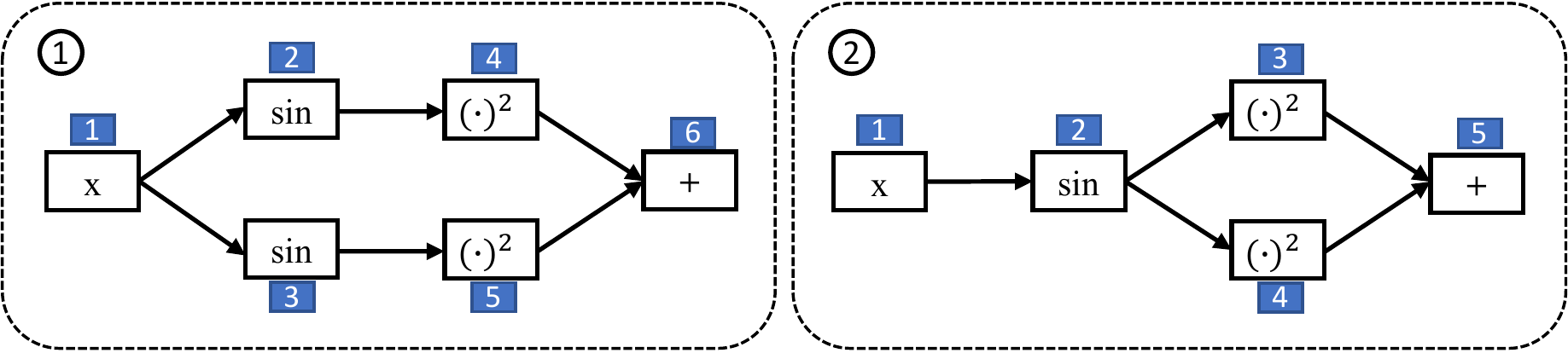}
\caption{\small Computations can be represented by \DAGs. Note that the left and right \DAGs represent the same computation.
}
\label{computationVSstructure}
\end{wrapfigure}

The set of elementary operations $\mathcal{O}$ depends on specific applications. For example, when we are interested in computations given by a calculator, $\mathcal{O}$ will be the set of all the operations defined on the functional buttons, such as $+$, $-$, $\times$, $\div$, etc. 
When modeling neural networks, $\mathcal{O}$ can be a predefined set of basic layers, such as 3$\times$3 convolution, 5$\times$5 convolution, 2$\times$2 max pooling, etc. A computation can be represented as a directed acyclic graph (\DAG), with directed edges representing signal flow directions among node operations. The graph must be acyclic, since otherwise the input signal will go through an infinite number of operations so that the computation never stops. Figure \ref{computationVSstructure} shows two examples. 
Note that the two different \DAGs in Figure \ref{computationVSstructure} represent the same computation, as the input signal goes through exactly the same operations. We discuss it further in Appendix \ref{appendix:difference}.



\subsection{Encoding}
We first introduce the encoder of \DVAE, 
which can be seen as a graph neural network (\GNN) using an asynchronous message passing scheme.
Given a \DAG $G$, we assume there is a single starting node which does not have any predecessors (e.g., the input layer of a neural architecture). If there are multiple such nodes, we add a virtual starting node connecting to all of them. 

Similar to standard \GNNs, we use an update function $\mathcal{U}$ to compute the hidden state of each node based on its neighbors' incoming message. The hidden state of node $v$ is given by:
\begin{equation}
\vec{h}_v = \mathcal{U}(\vec{x}_v, \vec{h}_v^{\text{in}}),
\label{update}
\end{equation}
where $\vec{x}_v$ is the one-hot encoding of $v$'s type, and $\vec{h}_v^{\text{in}}$ represents the incoming message to $v$. $\vec{h}_v^{\text{in}}$ is given by aggregating the hidden states of $v$'s predecessors using an aggregation function $\mathcal{A}$:
\begin{equation}
    \vec{h}_v^{\text{in}} = \mathcal{A}(\big\{\vec{h}_u: u\rightarrow v \big\}),
    \label{aggregate}
\end{equation}
where $u \rightarrow v$ denotes there is a directed edge from $u$ to $v$, and $\big\{\vec{h}_u: u\rightarrow v \big\}$ represents a multiset of $v$'s predecessors' hidden states. If an empty set is input to $\mathcal{A}$ (corresponding to the case for the starting node without any predecessors), we let $\mathcal{A}$ output an all-zero vector.

\begin{figure}[tp]
\centering
\includegraphics[width=0.95\textwidth]{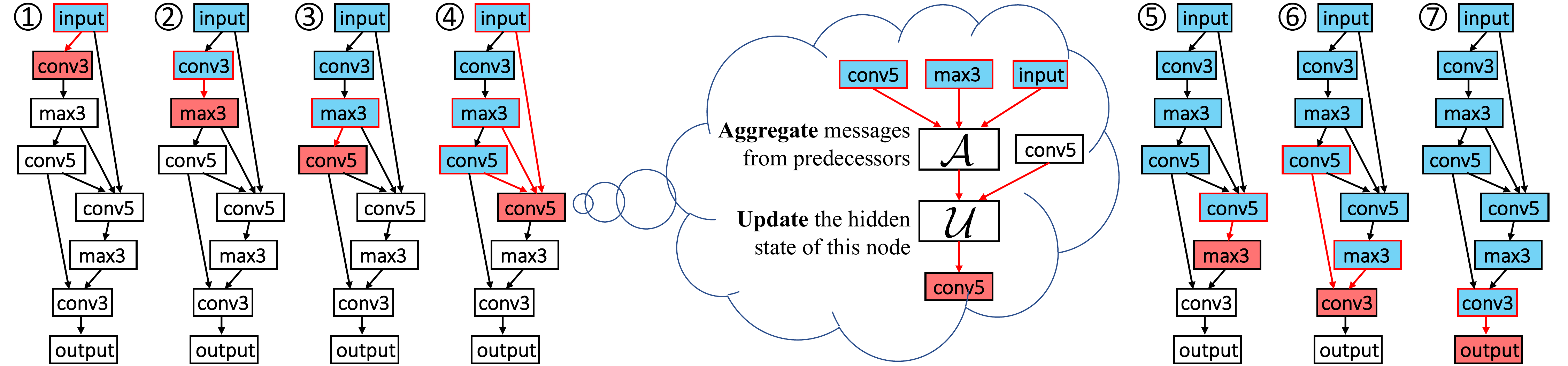}
\caption{\small An illustration of the encoding procedure for a neural architecture. Following a topological ordering, we iteratively compute the hidden state for each node (red) by feeding in its predecessors' hidden states (blue). This simulates how an input signal goes through a computation, with $\vec{h}_v$ simulating the output signal at node $v$. 
}
\label{encoding}
\end{figure}

Compared to the traditional simultaneous message passing, in \DVAE the message passing for a node must wait until all of its predecessors' hidden states have already been computed. 
This simulates how a computation is really performed -- to execute some operation, we also need to wait until all its input signals are ready. So how to make sure all the predecessor states are available when a new node comes? One solution is that we can sequentially perform message passing for nodes following a \textit{topological ordering} of the \DAG. We illustrate this encoding process in Figure \ref{encoding}.


After all nodes' hidden states are computed, we use $\vec{h}_{v_n}$, the hidden state of the ending node $v_n$ without any successors, as the output of the encoder. 
Then we feed $\vec{h}_{v_n}$ to two \MLPs to get the mean and variance parameters of the posterior approximation $q_\phi(\vec{z}|G)$ in (\ref{VAE_loss}). 
If there are multiple nodes without successors, we again add a virtual ending node connecting from all of them.

Note that although topological orderings are usually not unique for a \DAG, we can take any one of them as the message passing order while ensuring the encoder output is always the same, revealed by the following theorem. We include all theorem proofs in the appendix.

\begin{theorem}
The \DVAE encoder is invariant to node permutations of the input \DAG if the aggregation function $\mathcal{A}$ is invariant to the order of its inputs. 
\label{perm_inv}
\end{theorem}
Theorem \ref{perm_inv} means isomorphic \DAGs are always encoded the same, no matter how we index the nodes. 
It also indicates that so long as we encode a \DAG complying with its partial order, we can perform message passing in arbitrary order (even parallelly for some nodes) with the same encoding result. 

The next theorem shows another property of \DVAE that is crucial for its success in modeling \DAGs, i.e., it is able to injectively encode computations on \DAGs.

\begin{theorem}
Let $G$ be any \DAG representing some computation $C$. Let $v_1,\ldots, v_n$ be its nodes following a topological order each representing some operation $o_i, 1\leq i \leq n$, where $v_n$ is the ending node. Then, the encoder of \DVAE maps $C$ to $\vec{h}_{v_n}$ injectively if $\mathcal{A}$ is injective and $\mathcal{U}$ is injective.
\label{injective}
\end{theorem}
The significance of Theorem \ref{injective} is that it provides a way to injectively encode computations on \DAGs, so that every computation has a unique embedding in the latent space. 
Therefore, instead of performing optimization in the original discrete space, we may alternatively perform optimization in the \textbf{continuous latent space}. 
In this well-behaved Euclidean space, distance is well defined, and principled Bayesian optimization can be applied to search for latent points with high performance scores, which transforms the discrete optimization problem into an easier continuous problem. 


Note that Theorem \ref{injective} states \DVAE injectively encodes computations on graph structures, rather than graph structures themselves. 
Being able to injectively encode graph structures is a very strong condition, as it implies an efficient algorithm to solve the challenging graph isomorphism (\acro{GI}) problem. 
Luckily, here what we really care about are computations instead of structures, since we do not want to differentiate two different structures $G_1$ and $G_2$ as long as they represent the \textbf{same computation}. 
Figure \ref{computationVSstructure} shows such an example. 
Our \DVAE can identify that the two \DAGs in Figure~\ref{computationVSstructure} actually represent the same computation by encoding them to the same vector, 
while those encoders focusing on encoding structures might fail to capture the underlying computation and output different vectors. 
We discuss more advantages of Theorem \ref{injective}  in optimizing \DAGs in Appendix \ref{advantages}. 

To model and learn the injective functions $\mathcal{A}$ and $\mathcal{U}$, we resort to neural networks thanks to the universal approximation theorem \cite{hornik1989multilayer}. 
For example, we can let $\mathcal{A}$ be a gated sum:
\begin{align}
    \vec{h}_v^{\text{in}} = \sum\nolimits_{u\rightarrow v} g(\vec{h}_u) \odot m(\vec{h}_u),
    \label{aggregate2}
\end{align}
where $m$ is a mapping network and $g$ is a gating network. 
Such a gated sum can model injective multiset functions \cite{xu2018powerful}, and is invariant to input order.
To model the injective update function $\mathcal{U}$, we can use a gated recurrent unit (\GRU) \cite{cho2014learning}, with $\vec{h}_v^{\text{in}}$ treated as the input hidden state:
\begin{align}
\vec{h}_v = \mathrm{GRU}_e(\vec{x}_v, \vec{h}_v^{\text{in}}).
\label{update2}
\end{align}
Here the subscript $e$ denotes ``encoding''. Using a \GRU also allows reducing our framework to traditional sequence to sequence modeling frameworks \cite{sutskever2014sequence}, as discussed in \ref{discussion}.

The above aggregation and update functions can be used to encode general computation graphs. For neural architectures, depending on how the outputs of multiple previous layers are aggregated as the input to a next layer, we will make a modification to (\ref{aggregate2}), which is discussed in Appendix~\ref{encodeNN}. For Bayesian networks, we also make some modifications to their encoding due to the special d-separation properties of Bayesian networks, which is discussed in Appendix~\ref{encodeBN}.

\begin{figure}[tp]
\centering
\includegraphics[width=0.99\textwidth]{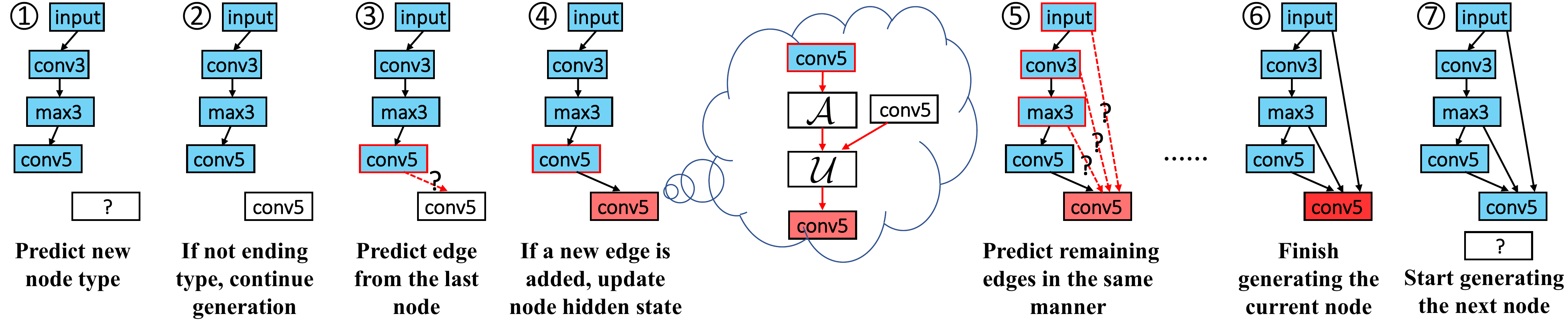}
\caption{\small An illustration of the steps for generating a new node.}
\label{decoding}
\end{figure}

\subsection{Decoding}\label{sec:decoding}
We now describe how \DVAE decodes latent vectors to \DAGs (the generative part). 
The \DVAE decoder uses the same asynchronous message passing scheme as in the encoder to learn intermediate node and graph states. 
Similar to (\ref{update2}), the decoder uses another \GRU, denoted by $\mathrm{GRU}_d$, to update node hidden states during the generation. 
Given the latent vector $\vec{z}$ to decode, we first use an \MLP to map $\vec{z}$ to $\vec{h_0}$ as the initial hidden state to be fed to $\mathrm{GRU}_d$. Then, the decoder constructs a \DAG node by node. 
For the $i^{\text{th}}$ generated node $v_i$, the following steps are performed:

\begin{enumerate}[leftmargin=*,itemsep=-1pt,topsep=-1pt]
\item Compute $v_i$'s type distribution using an \MLP $f_{\text{add\_vertex}}$ (followed by a softmax) based on the current graph state $\vec{h}_G:=\vec{h}_{v_{i-1}}$.
\item Sample $v_i$'s type. If the sampled type is the ending type, stop the decoding, connect all loose ends (nodes without successors) to $v_i$, and output the \DAG; otherwise, continue the generation.
\item Update $v_i$'s hidden state by $\vec{h}_{v_i} = \mathrm{GRU}_d(\vec{x}_{v_i}, \vec{h}_{v_i}^{\text{in}})$, where $\vec{h}_{v_i}^{\text{in}}=\vec{h_0}$ if $i=1$; otherwise, $\vec{h}_{v_i}^{\text{in}}$ is the aggregated message from its predecessors' hidden states given by equation (\ref{aggregate2}).
\item For $j \!=\! i\!-\!1, i\!-\!2,\ldots, 1$: (a) compute the edge probability of $(v_j,v_i)$ using an \MLP $f_{\text{add\_edge}}$ based on $\vec{h}_{v_j}$ and $\vec{h}_{v_i}$; (b) sample the edge; and (c) if a new edge is added, update $\vec{h}_{v_i}$ using step 3.
\end{enumerate}

The above steps are iteratively applied to each new generated node, until step 2 samples the ending type. 
For every new node, we first predict its node type based on the current graph state, and then sequentially predict whether each existing node has a directed edge to it based on the existing and current nodes' hidden states. 
Figure \ref{decoding} illustrates this process. Since edges always point to new nodes, the generated graph is guaranteed to be acyclic. 
Note that we maintain hidden states for both the current node and existing nodes, and keep updating them during the generation. 
For example, whenever step 4 samples a new edge between $v_j$ and $v_i$, we will update $\vec{h}_{v_i}$ to reflect the change of its predecessors and thus the change of the computation so far. 
Then, we will use the new $\vec{h}_{v_i}$ for the next prediction. Such a dynamic updating scheme is flexible, computation-aware, and always uses the up-to-date state of each node to predict next steps. 
In contrast, methods based on \RNNs \cite{kusner2017grammar,you2018graphrnn} do not maintain states for old nodes, and only use the current \RNN state to predict the next step. 

In step 4, when sequentially predicting incoming edges from previous nodes, we choose the reversed order $i-1,\ldots,1$ instead of $1,\ldots,i-1$ or any other order. This is based on the prior knowledge that a new node $v_i$ is more likely to firstly connect from the node $v_{i-1}$ immediately before it. For example, in neural architecture design, when adding a new layer, we often first connect it from the last added layer, and then decide whether there should be skip connections from other previous layers. Note that however, such an order is not fixed and can be flexible according to specific applications. 

\subsection{Training}
During the training phase, we use teacher forcing \cite{jin2018junction} to measure the reconstruction loss: following the topological order with which the input \DAG's nodes are consumed, we sum the negative log-likelihood of each decoding step by forcing them to generate the ground truth node type or edge at each step. This ensures that the model makes predictions based on the correct histories. Then, we optimize the \VAE loss (the negative of (\ref{VAE_loss})) using mini-batch gradient descent following \cite{jin2018junction}. 
Note that teacher forcing is only used in training. During generation, we sample a node type or edge at each step according to the decoding distributions described in Section \ref{sec:decoding} and calculate subsequent decoding distributions based on the sampled results.

\subsection{Discussion and model extensions}\label{discussion}

\noindent \textbf{Relation with RNNs.}
The \DVAE encoder and decoder can be reduced to ordinary \RNNs when the input \DAG is reduced to a chain of nodes. 
Although we propose \DVAE from a \GNN's perspective, our model can also be seen as a generalization of traditional sequence modeling frameworks \cite{sutskever2014sequence,bowman2015generating} where a timestamp depends only on the timestamp immediately before it, 
to the \DAG case where a timestamp has multiple previous dependencies. 
As special \DAGs, similar ideas have been explored for trees \cite{tai2015improved,jin2018junction}, where a node can have multiple incoming edges yet only one outgoing edge.

\noindent \textbf{Bidirectional encoding.} \DVAE's encoding process can be seen as simulating how an input signal goes through a \DAG, with $\vec{h}_v$ simulating the output signal at each node $v$. 
This is also known as \textit{forward propagation} in neural networks. Inspired by the bidirectional \RNN \cite{schuster1997bidirectional}, we can also use another \GRU to reversely encode a \DAG (i.e., reverse all edge directions and encode the \DAG again), thus simulating the \textit{backward propagation} too. 
After reverse encoding, we get two ending states, which are concatenated and linearly mapped to their original size as the final output state. 
We find this bidirectional encoding can increase the performance and convergence speed on neural architectures. 

\noindent \textbf{Incorporating vertex semantics.} 
Note that \DVAE currently uses one-hot encoding of node types as $\vec{x}_v$, which does not consider the semantic meanings of different node types. For example, a $3\times 3$ convolution layer might be functionally very similar to a $5\times 5$ convolution layer, while being functionally distinct from a max pooling layer. We expect incorporating such semantic meanings of node types to be able to further improve \DVAE's performance. For example, we can use pretrained embeddings of node types to replace the one-hot encoding. We leave it for future work.

\section{Experiments}
\label{sec:experiments}

We validate the proposed \DAG variational autoencoder (\DVAE) on two \DAG optimization tasks: 

\begin{itemize}[leftmargin=*,itemsep=-1pt,topsep=-1pt]
\item \textbf{Neural architecture search.}
Our neural network dataset contains 19,020 neural architectures from the \ENAS software \cite{pham2018efficient}. Each neural architecture has 6 layers (excluding input and output layers) sampled from: $3\times 3$ and $5\times 5$ convolutions, $3\times 3$ and $5\times 5$ depthwise-separable convolutions \cite{chollet2017xception}, $3\times 3$ max pooling, and $3\times 3$ average pooling. We evaluate each neural architecture's weight-sharing accuracy \cite{pham2018efficient} (a proxy of the true accuracy) on CIFAR-10 \cite{krizhevsky2009learning} as its performance measure. We split the dataset into 90\% training and 10\% held-out test sets. We use the training set for \VAE training, and use the test set only for evaluation.

\item \textbf{Bayesian network structure learning.}
Our Bayesian network dataset contains 200,000 random 8-node Bayesian networks from the \texttt{bnlearn} package \citep{JSSv035i03} in R. For each network, we compute the Bayesian Information Criterion (\BIC) score to measure the performance of the network structure for fitting the Asia dataset \citep{lauritzen1988local}. We split the Bayesian networks into 90\% training and 10\% test sets. For more details, please refer to Appendix \ref{asia}.
\end{itemize}

Following \cite{kusner2017grammar}, we do four experiments for each task:
\begin{itemize}[leftmargin=*,itemsep=-1pt,topsep=-1pt]
\item \textbf{Basic abilities of VAE models.} 
In this experiment, we perform standard tests to evaluate the reconstructive and generative abilities of a \VAE model for \DAGs, including reconstruction accuracy, prior validity, uniqueness and novelty. 
\item \textbf{Predictive performance of latent representation.} 
We test how well we can use the latent embeddings of neural architectures and Bayesian networks to predict their performances.
\item \textbf{Bayesian optimization.} 
This is the motivating application of \DVAE. 
We test how well the learned latent space can be used for searching for high-performance \DAGs through Bayesian optimization.
\item \textbf{Latent space visualization.} 
We visualize the latent space to qualitatively evaluate its smoothness.
\end{itemize}

Since there is little previous work on \DAG generation, we compare \DVAE with four generative baselines adapted for \DAGs: \SVAE, \GraphRNN, \GCN and \dgmg. Among them, \SVAE \cite{bowman2015generating} and \GraphRNN \cite{you2018graphrnn} are adjacency-matrix-based methods; \GCN \cite{kipf2016semi} and \dgmg \cite{li2018learning} are graph-based methods which use simultaneous message passing to embed \DAGs. 
We include more details about these baselines and discuss \DVAE's advantages over them in Appendix \ref{baselines}. 
The training details are in Appendix \ref{appendix:training}. 
All the code and data 
are available at \url{https://github.com/muhanzhang/D-VAE}.
\subsection{Reconstruction accuracy, prior validity, uniqueness and novelty}\label{exp1}
Being able to accurately reconstruct input examples and generate valid new examples are basic requirements for \VAE models. In this experiment, we evaluate the models by measuring 1) how often they can reconstruct input \DAGs perfectly (Accuracy), 2) how often they can generate valid neural architectures or Bayesian networks from the prior distribution (Validity), 3) the proportion of unique \DAGs out of the valid generations (Uniqueness), and 4) the proportion of valid generations that are never seen in the training set (Novelty). 

We first evaluate each model's reconstruction accuracy on the test sets. Following previous work \cite{kusner2017grammar,jin2018junction}, we regard the encoding as a stochastic process. That is, after getting the mean and variance parameters of the posterior approximation $q_\phi(\vec{z}|G)$, we sample a $\vec{z}$ from it as $G$'s latent vector. To estimate the reconstruction accuracy, we sample $\vec{z}$ 10 times for each $G$, and decode each $\vec{z}$ 10 times too. Then we report the average proportion of the 100 decoded \DAGs that are identical to the input.
To calculate prior validity, we sample 1,000 latent vectors $\vec{z}$ from the prior distribution $p(\vec{z})$ 
and decode each latent vector 10 times. Then we report the proportion of valid \DAGs in these 10,000 generations. A generated \DAG is valid if it can be read by the original software which generated the training data. More details about the validity experiment are in Appendix \ref{validity}.

\begin{table}[t]
\centering
\caption{\small Reconstruction accuracy, prior validity, uniqueness and novelty (\%).}
\resizebox{0.86\textwidth}{!}{
\begin{tabular}{@{}ccccccccc@{}}
\toprule
        & \multicolumn{4}{c}{Neural architectures}           & \multicolumn{4}{c}{Bayesian networks}        \\ \cmidrule(r{0.5em}){2-5} \cmidrule(l{0.5em}){6-9}
Methods & Accuracy   & Validity    & Uniqueness  & Novelty     & Accuracy  & Validity  & Uniqueness  & Novelty   \\ \midrule
D-VAE      & 99.96   & 100.00   & 37.26   & 100.00    & 99.94     & 98.84  & 38.98   & 98.01          \\
S-VAE      & 99.98   & 100.00   & 37.03   & 99.99     & 99.99     & 100.00 & 35.51   & 99.70          \\
GraphRNN   & 99.85   & 99.84    & 29.77   & 100.00    & 96.71     & 100.00 & 27.30   & 98.57  \\
GCN        & 98.70    & 99.53    & 34.00   & 100.00    & 99.81     & 99.02  & 32.84   & 99.40   \\
DeepGMG    & 94.98   & 98.66    & 46.37   & 99.93   &47.74  &98.86  &57.27  &98.49\\
\bottomrule
\end{tabular}
}
\label{results1}
\end{table}

We show the results in Table \ref{results1}. Among all the models, \DVAE and \SVAE generally perform the best. We find that \DVAE, \SVAE and \GraphRNN all have near perfect reconstruction accuracy, prior validity and novelty. However, \DVAE and \SVAE show higher uniqueness, meaning that they generate more diverse examples. 
\GCN and \dgmg have worse reconstruction accuracies for neural architectures due to nonzero training losses. This is because the simultaneous message passing scheme in them focus more on learning local graph structures, but fail to encode the computation represented by the entire neural network. Besides, the sum pooling after the message passing might also lose some global topology information which is important for the reconstruction. The nonzero training loss of \dgmg acts like an early stopping regularizer, making \dgmg generate more unique graphs. Nevertheless, reconstruction accuracy is much more important than uniqueness in our tasks, since we want our embeddings to accurately remap to their original structures after latent space optimization. 


\subsection{Predictive performance of latent representation.}\label{predictive}
In this experiment, we evaluate how well the learned latent embeddings can predict the corresponding \DAGs' performances, which tests a \VAE's unsupervised representation learning ability. 
Being able to accurately predict a latent point's performance also makes it much easier to search for high-performance points in this latent space. 
Thus, the experiment is also an indirect way to evaluate a \VAE latent space's amenability for \DAG optimization. 
Following \cite{kusner2017grammar}, we train a sparse Gaussian process (\SGP)  model \cite{snelson2006sparse} with 500 inducing points on the embeddings of training data to predict the performance of unseen test data. We include the \SGP training details in Appendix \ref{sgp}.

\begin{wraptable}{L}{0.61\textwidth}
\centering
\vspace{-10pt}
\caption{\small Predictive performance of encoded means.}
\resizebox{0.61\textwidth}{!}{
\begin{tabular}{@{}ccccccc@{}}
\toprule
        &\multicolumn{2}{c}{Neural architectures}            & \multicolumn{2}{c}{Bayesian networks}        \\ \cmidrule(r{0.5em}){2-3} \cmidrule(l{0.5em}){4-5}
Methods &  RMSE         & Pearson's $r$  & RMSE  & Pearson's $r$  \\ \midrule
D-VAE   & \textbf{0.384$\pm$0.002} & \textbf{0.920$\pm$0.001}  & \textbf{0.300$\pm$0.004} & \textbf{0.959$\pm$0.001}        \\
S-VAE    & 0.478$\pm$0.002     & 0.873$\pm$0.001      &  0.369$\pm$0.003        &  0.933$\pm$0.001        \\ 
GraphRNN & 0.726$\pm$0.002     & 0.669$\pm$0.001      &  0.774$\pm$0.007        &  0.641$\pm$0.002      \\
GCN      & 0.485$\pm$0.006     & 0.870$\pm$0.001        &  0.557$\pm$0.006                & 0.836$\pm$0.002         \\
DeepGMG   & 0.433$\pm$0.002 &  0.897$\pm$0.001 & 0.788$\pm$0.007 & 0.625$\pm$0.002\\
\bottomrule
\end{tabular}
}
\label{results2}
\end{wraptable}

We use two metrics to evaluate the predictive performance of the latent embeddings (given by the mean of the posterior approximations $q_\phi(\vec{z}|G)$). One is the \RMSE between the \SGP predictions and the true performances. 
The other is the Pearson correlation coefficient (or Pearson's $r$), measuring how well the prediction and real performance tend to go up and down together. 
A small \RMSE and a large Pearson's $r$ indicate a better predictive performance. All the experiments are repeated 10 times and the means and standard deviations are reported.
Table \ref{results2} shows the results. We find that both the \RMSE and Pearson's $r$ of \DVAE are significantly better than those of the other models. A possible explanation is that \DVAE encodes the computation, while a \DAG's performance is primarily determined by its computation. Therefore, \DVAE's latent embeddings are more informative about performance. 
In comparison, adjacency-matrix-based methods (\SVAE and \GraphRNN) and graph-based methods with simultaneous message passing (\GCN and \dgmg) both only encode (local) graph structures without specifically modeling computations on \DAG structures. 
The better predictive power of \DVAE favors using a predictive model in its latent space to guide the search for high performance graphs.

\begin{figure*}[tp]
\centering
\includegraphics[width=0.99\textwidth]{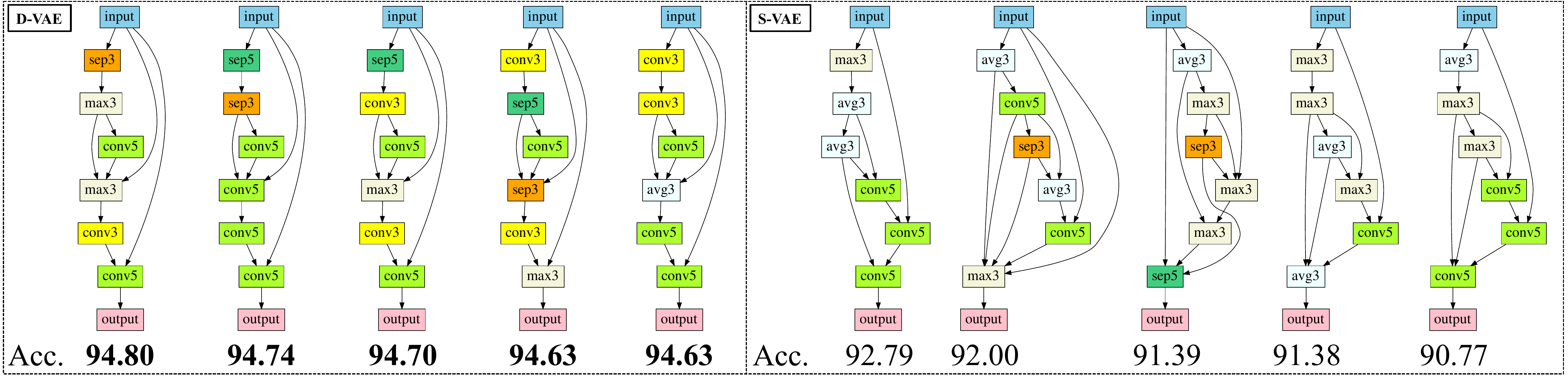}
\caption{\small Top 5 neural architectures found by each model and their true test accuracies. 
}
\label{top5}
\end{figure*}

\begin{figure}[t]
\centering
\includegraphics[width=1\textwidth]{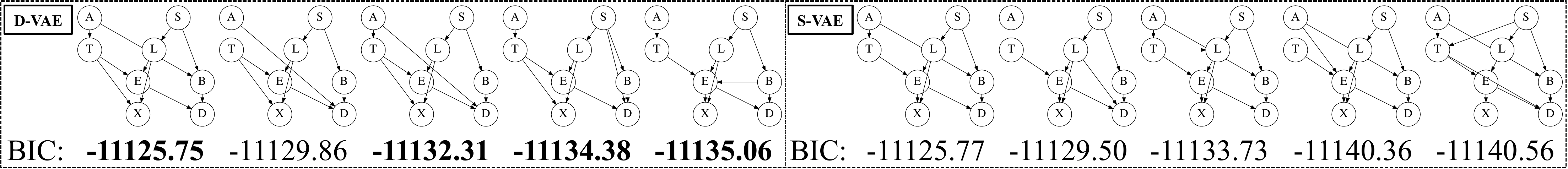}
\caption{\small Top 5 Bayesian networks found by each model and their \BIC scores (higher the better).}
\label{top5_BN}
\end{figure}

\begin{figure*}[tp]
\centering
\includegraphics[width=0.98\textwidth]{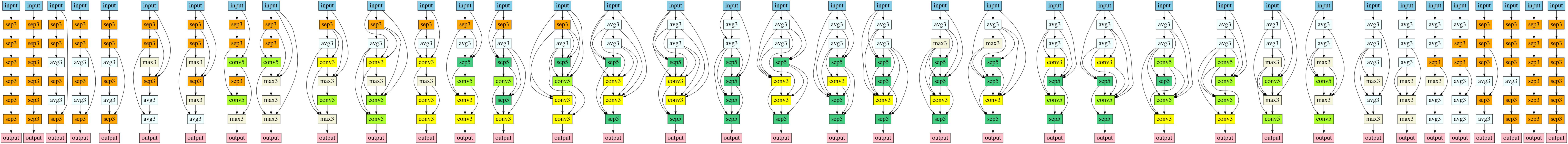}\\
\vspace{0.2cm}
\includegraphics[width=0.98\textwidth]{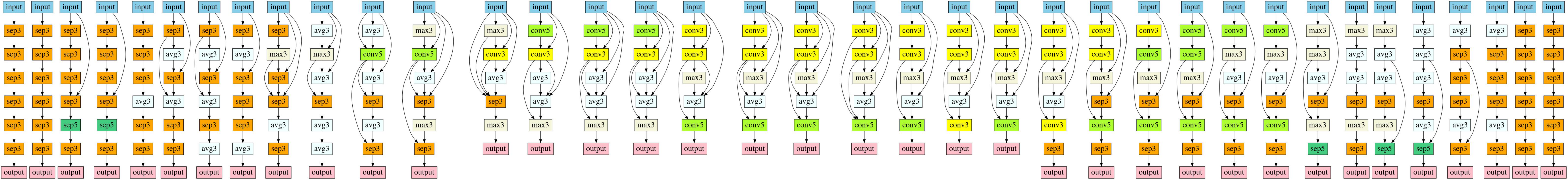}
\caption{\small Great circle interpolation starting from a point and returning to itself. Upper: \DVAE. Lower: \SVAE.}
\label{NN_interpolate}
\end{figure*}



\subsection{Bayesian optimization}\label{exp:bo}
We perform Bayesian optimization (\bo) using the two best models, 
\DVAE and \SVAE, validated by previous experiments. 
Based on the \SGP model from the last experiment, we perform 10 iterations of batch \bo, and average results across 10 trials. 
Following \citet{kusner2017grammar}, in each iteration, a batch of 50 points are proposed by sequentially maximizing the expected improvement (\acro{EI}) acquisition function, using Kriging Believer \citep{ginsbourger2010kriging} to assume labels for previously chosen points in the batch. 
For each batch of selected points, we evaluate their decoded \DAGs' real performances and add them back to the \SGP to select the next batch. 
Finally, we check the best-performing \DAGs found by each model to evaluate its \DAG optimization performance. 


\noindent \textbf{Neural architectures.} For neural architectures, we select the top 15 found architectures in terms of their weight-sharing accuracies, and fully train them on CIFAR-10's train set to evaluate their true test accuracies. More details can be found in Appendix \ref{enas}. 
We show the 5 architectures with the highest true test accuracies in Figure \ref{top5}. As we can see, \DVAE in general found much better neural architectures than \SVAE. 
Among the selected architectures, \DVAE achieved a highest accuracy of 94.80\%, while \SVAE's highest accuracy was only 92.79\%. In addition, all the 5 architectures of \DVAE have accuracies higher than 94\%, indicating that \DVAE's latent space can stably find many high-performance architectures. More details about our \NAS experiments are in Appendix \ref{enas}.

\noindent \textbf{Bayesian networks.} We similarly report the top 5 Bayesian networks found by each model ranked by their \BIC scores in Figure \ref{top5_BN}. \DVAE generally found better Bayesian networks than \SVAE. The best Bayesian network found by \DVAE achieved a \BIC of -11125.75, which is better than the best network in the training set with a \BIC of -11141.89 (a higher \BIC score is better). Note that \BIC is in log scale, thus the probability of our found network to explain the data is actually 1E7 times larger than that of the best training network. For reference, the true Bayesian network used to generate the Asia data has a \BIC of -11109.74. Although we did not exactly find the true network, our found network was close to it and outperformed all 180,000 training networks. Our experiments show that searching in an embedding space is a promising direction for Bayesian network structure learning.

\subsection{Latent space visualization}\label{visualization}
In this experiment, we visualize the latent spaces of the \VAE models to get a sense of their smoothness.

For neural architectures, we visualize the decoded architectures from points along a great circle in the latent space \citep{white2016sampling} (slerp). 
We start from the latent embedding of a straight network without skip connections. Imagine this latent embedding as a point on the surface of a sphere (visualize the earth). We randomly pick a great circle starting from this point and returning to itself around the sphere. Along this circle, we evenly pick 35 points and visualize their decoded neural architectures in Figure \ref{NN_interpolate}. As we can see, both \DVAE and \SVAE show relatively smooth interpolations by changing only a few node types or edges each time. Visually speaking, \SVAE's structural changes are even smoother. This is because \SVAE treats \DAGs as strings, thus tending to embed \DAGs with few differences in string representations to similar regions of the latent space without considering their computational differences (see Appendix \ref{baselines} for more discussion of this problem). In contrast, \DVAE models computations, and focuses more on the smoothness w.r.t. computation rather than structure.



\begin{wrapfigure}[11]{L}{0.6\textwidth}
\centering
\vspace{-10pt}
\includegraphics[width=0.57\textwidth]{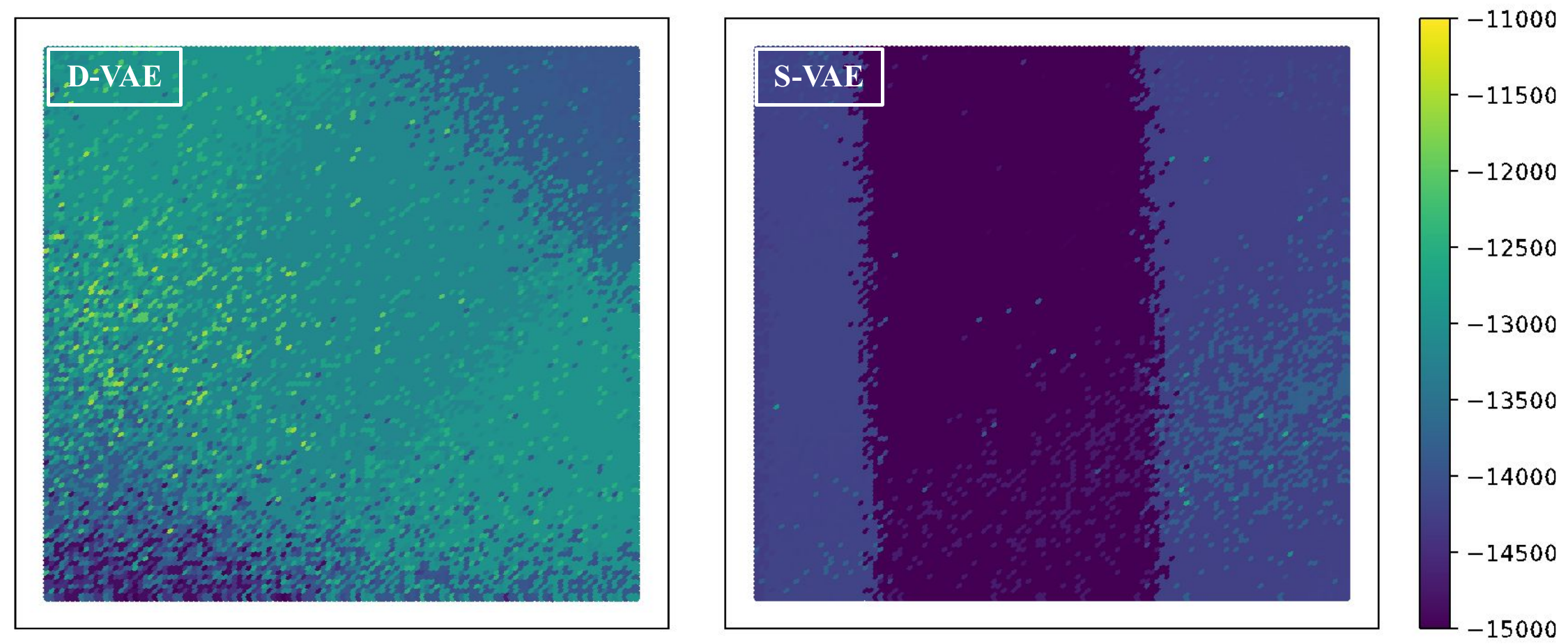}
\caption{\small Visualizing a principal 2-D subspace of the latent space.}
\label{BN_vis}
\end{wrapfigure}

For Bayesian networks, we directly visualize the \BIC score distribution of the latent space. To do so, we reduce its dimensionality by choosing a 2-D subspace spanned by the first two principal components of the training data's embeddings. In this low-dimensional subspace, we compute the \BIC scores of all the points evenly spaced within a $[-0.3,0.3]$ grid and visualize the scores using a colormap in Figure \ref{BN_vis}. As we can see, \DVAE seems to better differentiate high-score points from low-score ones and shows more smoothly changing \BIC scores. In comparison, \SVAE shows sharp boundaries and seems to mix high-score and low-score points more severely. The smoother latent space might be the key reason for the better Bayesian optimization performance with \DVAE. Furthermore, we notice that \DVAE's 2-D latent space is brighter; one explanation is the two principal components of \DVAE explain more variance (59\%) of training data than those of \SVAE (17\%). Thus, along the two principal components of \SVAE we will see less points from the training distribution. These out-of-distribution points tend to decode to not very good Bayesian networks, thus are darker. This also indicates that \DVAE learns a more compact latent space.

\section{Conclusion}
In this paper, we have proposed \DVAE, a \GNN-based deep generative model for \DAGs. 
\DVAE uses a novel asynchronous message passing scheme to encode a \DAG respecting its partial order, which explicitly models the computations on \DAGs.
By performing Bayesian optimization in \DVAE's latent spaces, we offer promising new directions to two important problems, neural architecture search and Bayesian network structure learning. We hope \DVAE can inspire more research on studying \DAGs and their applications in the real world. 



\label{sec:conclusion}

\subsubsection*{Acknowledgments}
The work is supported in part by the III-1526012 and SCH-1622678 grants from the National Science Foundation and grant 1R21HS024581 from the National Institute of Health. The authors would like to thank Liran Wang for the helpful discussions.

{\small
\bibliography{main}}

\newpage
\newpage
\onecolumn
\begin{appendices}

\section{More Related Work}\label{related}
Both neural architecture search (\NAS) and Bayesian network structure learning (\BNSL) are subfields of AutoML. See \citet{zoller2019survey} for a survey. We have given a brief overview of \NAS and \BNSL in Section \ref{main:related}. Below we discuss several works most related to our work in more detail.

\citet{luo2018neural} proposed a novel \NAS approach called Neural Architecture Optimization (\acro{NAO}). 
The basic idea is to jointly learn an encoder-decoder between networks and a \emph{continuous} space, and also a performance predictor $f$ that maps the continuous representation of a network to its performance on a given dataset; 
then they perform two or three iterations of gradient descent on $f$ to find better architectures in the continuous space, which are then decoded to real networks to evaluate. 
This methodology is similar to that of \citet{gomez2018automatic} and \citet{jin2018junction} for molecule optimization; also similar to \citet{mueller2017sequence} for slightly revising a sentence.

There are several key differences comparing to our approach. 
First, \acro{NAO} uses strings (e.g. ``node-2 conv 3x3 node1 max-pooling 3x3'') to represent neural architectures, whereas we directly use graph representations, which is more natural and generally applicable to other graphs such as Bayesian network structures.
Second, \acro{NAO} uses supervised learning instead of unsupervised learning, which means it needs to first evaluate a considerable amount of randomly sampled graphs on a typically large dataset (e.g. train many neural networks), and use these results to supervise the training of the autoencoder. Given a new dataset, the autoencoder needs to be completely retrained. In contrast, we train our variational autoencoder in a fully unsupervised manner, so the model is of general purposes. 


\citet{fusi2018probabilistic} proposed a novel AutoML algorithm also using model embedding, but with a matrix factorization approach. They first construct a matrix of performances of thousands of ML pipelines 
on hundreds of datasets; 
then they use a probabilistic matrix factorization to get the latent representations of the pipelines.
Given a new dataset, Bayesian optimization with the expected improvement heuristic is used to find the best pipeline. 
This approach only allows us to choose from predefined off-the-shelf ML models, hence its flexibility is somewhat limited. 

\citet{kandasamy2018neural} use Bayesian optimization for \NAS; they define a kernel that measures the similarities between networks by solving an optimal transport problem, and in each iteration, they use some evolutionary heuristics to generate a set of candidate networks based on making small modifications to existing networks, and use expected improvement to choose the next one to evaluate.
This work is similar to ours in terms of the application of Bayesian optimization. However, defining a kernel to measure the similarities between discrete structures is a non-trivial problem.
In addition, the discrete search space is heuristically extrapolated near existing architectures, which makes the search essentially local. In contrast, we directly fit a Gaussian process over the entire continuous latent space, enabling more global optimization.

Using Gaussian process (\gp) for Bayesian network structure learning has also been studied before. \citet{yackley2012smoothness} analyzed the smoothness of \acro{BD}e score,
showing that a local change (e.g. adding an edge) can change the score by at most $\mathcal{O}(\log n)$, where $n$ is the number of training points. They proposed to use \gp as a proxy for the score to accelerate the search. \citet{anderson2009fast} used \gp to model the \acro{BD}e score, and showed that the probability of improvement is better than that of using hill climbing to guide the local search. 
However, these methods still heuristically and locally operate in the discrete space, whereas our latent space makes both local and global methods such as gradient descent and Bayesian optimization applicable in a principled manner.


Recently, \citet{zheng2018dags} also proposed a continuous optimization approach for \BNSL, where the decision variable is the adjacency matrix of the \DAG and the objective function is least square loss based on linear structural equation modeling (\acro{SEM}); acyclicity is ensured by a novel equality constraint. \citet{yu2019dag} generalize this approach to nonlinear \acro{SEM} using \VAE. We highlight several key differences from our approach: 1) they directly optimize the adjacency matrix, but we optimize a learned latent representation of the \DAGs; 2) they ensure acyclicity by enforcing an equality constraint, but for our method acyclicity is automatically guaranteed by the decoding process; 3) They use gradient-based optimization, but we use global black-box optimization. 4) Their methods are specific to \BNSL, but ours applies to general \DAG optimization; 5) The usage of \VAE is totally different: they use \VAE as a generative model for the data (sampled from the \DAG), but we use \VAE as a generative model for \DAGs.

\section{Graph Structure vs. Computation vs. Function}\label{appendix:difference}
In Section \ref{sec:D-VAE} we defined computation. Here we discuss the differences among \DAG structure, computation and function. A \DAG structure with operations on nodes define a computation, and two \DAGs can define the same computation, which are illustrated in Figure \ref{computationVSstructure}. A computation defines a function, and two computations can define the same function. For example, computation $C_1 := x + 1 - 1$ defines a function $f(x) = x$, while computations $C_2 := x - 1 + 1$ and $C_3 := x$ also define the function $f(x) = x$. However, $C_1$, $C_2$ and $C_3$ are different computations. In other words, a computation is (informally speaking) a process which focuses on the course of how the input is processed into the output, while a function is a mapping from input to output which does not care about the process. 

Sometimes, the same computation can also define different functions, e.g., two identical neural architectures will represent different functions given they are trained differently (since the weights of their layers will be different). In \DVAE, we model computations instead of functions, since 1) modeling functions is much harder than modeling computations (requires understanding the semantic meaning of each operation, such as the cancelling out of $+$ and $-$), and 2) modeling functions additionally requires knowing the parameters of some operations, which are unknown before training.

Note also that in Definition \ref{computation}, we only allow one single input signal. But in real world a computation sometimes has multiple initial input signals. However, the case of multiple input signals can be reduced to the single input case by adding an initial assignment operation that assigns the combined input signal to their corresponding next-level operations. For ease of presentation, we uniformly assume single input throughout the paper.

\section{Proof of Theorem \ref{perm_inv}}
\begin{proof}
Let $v_1$ be the starting node with no predecessors. By assumption, $v_1$ is the single starting node no matter how we permute the nodes of the input \DAG. For $v_1$, the aggregation function $\mathcal{A}$ always outputs a zero vector. Thus, $\vec{h}_{v_1}^{\text{in}}$ is invariant to node permutations. Subsequently, the hidden state $\vec{h}_{v_1} = \mathcal{U}(\vec{x}_{v_1},\vec{h}_{v_1}^{\text{in}})$ is also invariant to node permutations.

Now we prove the theorem by structural induction. Consider node $v$. Suppose for every predecessor $u$ of $v$, the hidden state $\vec{h}_{u}$ is invariant to node permutations. We will show that $\vec{h}_{v}$ is also invariant to node permutations. Notice that in (\ref{aggregate}), the output $\vec{h}_v^{\text{in}}$ by $\mathcal{A}$ is invariant to node permutations, since $\mathcal{A}$ is invariant to the order of its inputs $\vec{h}_u$, and all $\vec{h}_u$ are invariant to node permutations. Subsequently, node $v$'s hidden state $\vec{h}_v = \mathcal{U}(\vec{x}_{v},\vec{h}_{v}^{\text{in}})$ is invariant to node permutations. By induction, we know that every node's hidden state is invariant to node permutations, including the ending node's hidden state. Thus, the \DVAE encoder is invariant to node permutations.
\end{proof}

\section{Proof of Theorem \ref{injective}}
\begin{proof}
Suppose there is an arbitrary input signal $x$ fed to the starting node $v_1$. For convenience, we will use $C_i(x)$ to denote the output signal at vertex $v_i$, where $C_i$ represents the composition of all the operations along the paths from $v_1$ to $v_i$. 

For the starting node $v_1$, remember we feed a fixed $\vec{h}_{v_1}^{\text{in}}=\mathbf{0}$ to (\ref{update}), thus $\vec{h}_{v_1}$ is also fixed. Since $C_1$ also represents a fixed input operation, we know that the mapping from $C_1$ to $\vec{h}_{v_1}$ is injective. Now we prove the theorem by induction. Assume the mapping from $C_j$ to $\vec{h}_{v_j}$ is injective for all $1\leq j < i$. We will prove that the mapping from $C_i$ to $\vec{h}_{v_i}$ is also injective.

Let $\phi_j(C_j) = \vec{h}_{v_j}$ where $\phi_j$ is injective. Consider the output signal $C_i(x)$, which is given by feeding $\big\{C_j(x): v_j\rightarrow v_i\big\}$ to $o_i$. Thus,
\begin{align}
C_i(x) &= o_i(\big\{C_j(x): v_j\rightarrow v_i\big\}).
\end{align}
In other words, we can write $C_i$ as
\begin{align}
C_i = \psi(o_i, \big\{ C_j: v_j\rightarrow v_i \big\}),
\label{psi}
\end{align}
where $\psi$ is an injective function used for defining the composite computation $C_i$ based upon $o_i$ and $\big\{ C_j: v_j\rightarrow v_i \big\}$. Note that $\big\{ C_j: v_j\rightarrow v_i \big\}$ can be either unordered or ordered depending on the operation $o_i$. For example, if $o_i$ is some symmetric operations such as adding or multiplication, then $\big\{ C_j: v_j\rightarrow v_i \big\}$ can be unordered. If $o_i$ is some operation like subtraction or division, then $\big\{ C_j: v_j\rightarrow v_i \big\}$ must be ordered.

With (\ref{update}) and (\ref{aggregate}), we can write the hidden state $\vec{h}_{v_i}$ as follows:
\begin{align}
\vec{h}_{v_i} &= \mathcal{U}(\vec{x}_{v_i}, \mathcal{A}(\big\{\vec{h}_{v_j}: v_j\rightarrow v_i \big\})) \nonumber\\
              &= \mathcal{U}(O(o_i), \mathcal{A}(\big\{\phi_j(C_j): v_j\rightarrow v_i \big\})),
\end{align}
where $O$ is the injective one-hot encoding function mapping $o_i$ to $\vec{x}_{v_i}$. In the above equation, $\mathcal{U}, O, \mathcal{A}, \phi_j$ are all injective. Since the composition of injective functions is injective, there exists an injective function $\varphi$ so that
\begin{align}
\vec{h}_{v_i} = \varphi(o_i, \big\{C_j: v_j\rightarrow v_i \big\}).
\end{align}
Then combining (\ref{psi}) we have:
\begin{align}
\vec{h}_{v_i} &= \varphi \circ \psi^{-1} \psi(o_i, \big\{C_j: v_j\rightarrow v_i \big\}) \nonumber \\
              &= \varphi \circ \psi^{-1}(C_i).
\end{align}
$\varphi \circ \psi^{-1}$ is injective since the composition of injective functions is injective. Thus, we have proved that the mapping from $C_i$ to $\vec{h}_{v_i}$ is injective.
\end{proof}



\section{Modifications for Encoding Neural Architectures}\label{encodeNN}
According to Theorem \ref{injective}, to ensure \DVAE injectively encodes computations, we need the aggregation function $\mathcal{A}$ to be injective. Remember $\mathcal{A}$ takes the multiset $\big\{\vec{h}_u: u\rightarrow v \big\})$ as input. If the order of its elements does not matter, then the gated sum in (\ref{aggregate2}) can model this injective multiset function without issues. However, if the order matters (i.e., permuting the elements of $\big\{\vec{h}_u: u\rightarrow v \big\}$ makes $\mathcal{A}$ output different results), we need a different aggregation function that can encode such orders.

Whether the order should matter for $\mathcal{A}$ depends on whether the input order matters for the operations $o$ (see the proof for Theorem \ref{injective} for more details). For example, if multiple previous layers' outputs are summed or averaged as the input to a next layer in the neural networks, then $\mathcal{A}$ can be modeled by the gated sum in (\ref{aggregate2}) as the order of inputs does not matter. However, if these outputs are concatenated as the next layer's input, then the order does matter. In our experiments, the neural architectures use the second way to aggregate outputs from previous layers. The order of concatenation depends on a global order of the layers in a neural architecture. For example, if layer-2 and layer-4's outputs are input to layer-5, then layer-2's output will be before layer-4's output in their concatenation.

Since the gated sum in (\ref{aggregate2}) can only handle the unordered case, we can slightly modify (\ref{aggregate2}) in order to make it order-aware thus more suitable for our neural architectures. Our scheme is as follows:
\begin{align}
    \vec{h}_v^{\text{in}} = \sum_{u\rightarrow v} g(\text{Concat}(\vec{h}_u, \vec{x}_{\text{uid}})) \odot m(\text{Concat}(\vec{h}_u, \vec{x}_{\text{uid}})),
    \label{aggregate3}
\end{align}
where $\vec{x}_{\text{uid}}$ is the one-hot encoding of layer $u$'s global ID (1,2,3,\ldots). Such an aggregation function respects the concatenation order of the layers. We empirically observed that this aggregation function can increase \DVAE's performance on neural architectures compared to the plain aggregation function (\ref{aggregate2}). However, even using (\ref{aggregate2}) still outperformed all baselines.

\section{Modifications for Encoding Bayesian Networks}\label{encodeBN}

\begin{figure}[h]
\centering
\includegraphics[width=0.6\textwidth]{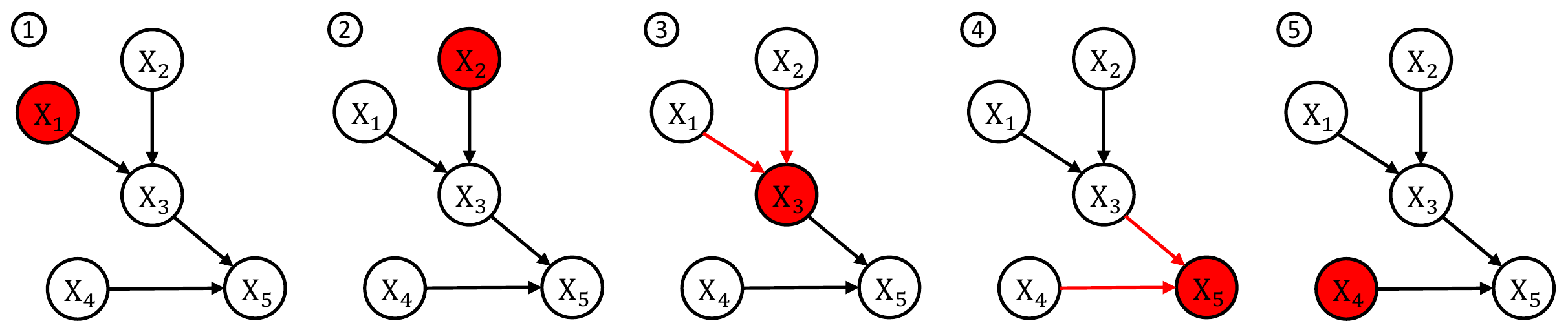}
\caption{\small An example Bayesian network and its encoding.}
\label{encodingBN}
\end{figure}

We also make some modifications when encoding Bayesian networks. One modification is that the aggregation function (\ref{aggregate2}) is changed to:
\begin{align}
    \vec{h}_v^{\text{in}} = \sum_{u\rightarrow v} g(\vec{x}_u) \odot m(\vec{x}_u).
    \label{aggregate_BN}
\end{align}
Compared to (\ref{aggregate2}), we replace $\vec{h}_u$ with the node type feature $\vec{x}_u$. This is due to the differences between computations on a neural architecture and on a Bayesian network. In a neural network, the signal flow follows the network architecture, where the output signal of a layer is fed as the input signals to its succeeding layers. Also in a neural network, what we are interested in is the result output by the final layer.
In contrast, for a Bayesian network, the graph represents a set of conditional dependencies among variables instead of a computational flow. 
In particular, for Bayesian network structure learning, we are often concerned about computing the (log) marginal likelihood score of a dataset given a graph structure, which is often decomposed into individual variables given their parents 
(see Definition 18.2 in \citet{koller2009probabilistic}). 
For example, in Figure \ref{encodingBN}, the overall score can be decomposed into 
$s(X_1) + s(X_2) + s(X_3\given X_1, X_2) + s(X_4) + s(X_5\given X_3, X_4)$.
To compute the score $s(X_5\given X_3, X_4)$ for $X_5$, we only need the values of $X_3$ and $X_4$; its grandparents $X_1$ and $X_2$ should have no influence on $X_5$.
Based on this intuition, when computing the hidden state of a node, we use the features $\vec{x}_u$ of its parents $u$ instead of $\vec{h}_u$, which ``d-separates'' the node from its grandparents. For the update function, we still use (\ref{update2}).

Also based on the decomposibility of the score, 
we make another modification for encoding Bayesian networks by using the sum of all node states as the final output state instead of only using the ending node state. Similarly, when decoding Bayesian networks, the graph state $\vec{h}_G := \sum_{j=1,\ldots, i-1} \vec{h}_{v_{j}}$.

Note that the combination of (\ref{aggregate_BN}) and (\ref{update2}) can injectively model the conditional dependence between $v$ and its parents $u$. In addition, using summing can model injective set functions \cite[Lemma~5]{xu2018powerful}. Therefore, the above encoding scheme is able to \textbf{injectively encode} the complete \textbf{conditional dependencies} of a Bayesian network, thus also the overall score function $s$ of the network. 

\section{Advantages of Encoding Computations in DAG Optimization}\label{advantages}
Here we discuss why \DVAE's ability to injectively encode computations (Theorem \ref{injective}) is of great benefit to performing \DAG optimization in the latent space. Firstly, our target is to find a \DAG that achieves high performance (e.g., accuracy of neural network, \BIC score of Bayesian network) on a given dataset. The performance of a \DAG is directly related to its computation. For example, given the same set of layer parameters, two neural networks with the same computation will have the same performance on a given test set. Since \DVAE encodes computations instead of structures, it allows \textbf{embedding \DAGs with similar performances to the same regions} in the latent space, rather than embedding \DAGs with merely similar structure patterns to the same regions. Subsequently, the latent space can be \textbf{smooth w.r.t.} \emph{performance} instead of \emph{structure.} Such smoothness can greatly facilitate searching for high-performance \DAGs in the latent space, since similar-performance \DAGs tend to locate near each other in the latent space instead of locating randomly, and modeling a smoothly-changing performance surface is much easier.

Note that Theorem \ref{injective} is a necessary condition for the latent space to be smooth w.r.t. performance, because if \DVAE cannot injectively encode computations, it might map two \DAGs representing completely different computations to the same encoding, making this point of the latent space arbitrarily unsmooth. Although there yet is no theoretical guarantee that the latent space must be smooth w.r.t. \DAGs' performances, we do empirically observe that the predictive performance and Bayesian optimization performance of \DVAE's latent space are significantly better than those of baselines, which is indirect evidence that \DVAE's latent space is smoother w.r.t. performance. Our visualization results also confirm the smoothness. See Section \ref{predictive}, \ref{exp:bo}, \ref{visualization} for details.

\section{More Details about Neural Architecture Search}\label{enas}

We use the efficient neural architecture search (\ENAS)'s software \cite{pham2018efficient} to generate the training and testing neural architectures. With these seed architectures, we can train a \VAE model and thus search for new high-performance architectures in the latent space.

\ENAS alternately trains two components: 1) an \RNN-based controller which is used to propose new architectures, and 2) the shared weights of the proposed architectures. It uses a weight-sharing scheme to obtain a quick but rough estimate of how good an architecture is. That is, it forces all the proposed architectures to use the same set of shared weights, instead of fully training each neural network individually. It assumes that an architecture with a high validation accuracy using the shared weights (i.e., the weight-sharing accuracy) is more likely to have a high test accuracy after fully retraining its weights from scratch.

We first run \ENAS in the macro space (Section 2.3 of \citep{pham2018efficient}) for 1000 epochs with 20 architectures proposed in each epoch. For all the proposed architectures excluding the first 1000 burn-in ones, we evaluate their weight-sharing accuracies using the shared weights from the last epoch. We further split the data into 90\% training and 10\% held-out test sets. Then our task becomes to train a \VAE on the training neural architectures, and then generate new high-performance architectures from the latent space based on Bayesian optimization. Note that our target performance measure here is the weight-sharing accuracy, not the true validation/test accuracy after fully retraining the architecture. This is because the weight-sharing accuracy takes around 0.5 second to evaluate, while fully training a network takes over 12 hours. In consideration of our limited computational resources, we choose the weight-sharing accuracy as our optimization target in the Bayesian optimization experiments.

After the Bayesian optimization finds a final set of architectures with high weight-sharing accuracies, we will fully train them to evaluate their true test accuracies on CIFAR-10. To fully train an architecture, we follow the original setting of \citep{pham2018efficient} to train each architecture on CIFAR-10's training set for 310 epochs, and report the last epoch's net's test accuracy. See \cite[Section~3.2]{pham2018efficient} for details.


Due to our constrained computational resources, we choose not to perform Bayesian optimization to optimize the true validation accuracy (obtained by fully training a neural network), which would be a more principled way for searching neural architectures. Nevertheless, we describe its procedure here for future explorations: After training the \DVAE, we have no architectures at all to initialize a Gaussian process regression on the true validation accuracy. Thus, we need to randomly pick up some points in the latent space, decode them into neural architectures, and get their true validation accuracies after full training. Then with these initial points, we start the Bayesian optimization similarly to Section \ref{exp:bo}, with the optimization target replaced by the true validation accuracy. Finally, we will find a set of architectures with the highest true validation accuracies, and report their true test accuracies. This experiment will take much longer time (possibly months of GPU time). Thus, it is very necessary to train multiple models parallelly on many machines, like \citep{zoph2016neural} does.

One might wonder why we train another generative model after we already have \ENAS. Firstly, \ENAS is a task-specific supervised model. It leverages the validation accuracy signals of the target task to guide the generation of new architectures based on reinforcement learning. For any new \NAS task, \ENAS needs to be completely retrained. In contrast, \DVAE is unsupervised. Once trained, it can be applied to \NAS tasks targeting different datasets. For example, although we use the neural architectures generated by the \ENAS targeting CIFAR-10 to train our \DVAE, once trained, we can use \DVAE's latent space to search neural architectures suitable for CV tasks other than CIFAR-10. In contrast, the trained \ENAS is not applicable to other tasks since it uses supervised signals from CIFAR-10. In other words, \ENAS is only used to generate a set of seed architectures for training \DVAE, and is not necessary for \DVAE. For example, we may also train \DVAE using the recent NAS-Bench-101 dataset\footnote{https://github.com/google-research/nasbench}, which we leave for future work.
Another exclusive advantage of \DVAE is that it provides a way to learn neural architecture embeddings, which can be used for downstream tasks such as visualization and classification, etc. 

In the Bayesian optimization experiments (Section \ref{exp:bo}), the best architecture found by \DVAE achieves a test accuracy of 94.80\% on CIFAR-10. Although not outperforming state-of-the-art \NAS techniques such as \acro{NAONet} which has an error rate of 2.11\%, our architecture only contains 3 million parameters compared to \acro{NAONet} + \acro{Cutout} which has 128 million parameters \cite{luo2018neural}. In addition, the search space is different between the two approaches: we directly search 6-layer \CNNs, while \acro{NAONet} searches 5-layer \CNN cells and stacks the found cell for 6 times to construct a \CNN, thus having much deeper final networks. Finally, \acro{NAONet} used 200 GPUs to fully train 1,000 architectures for 1 day, and added 4 times more filters after optimization. In comparison, we only used 1 GPU to evaluate the weight-sharing accuracy, and did not add filters to boost the performance. We emphasize that the main purpose of the paper is to introduce a \DAG generative model that is capable of \DAG optimization, rather than to break \NAS records. 


\section{More Details about Bayesian Network Structure Learning}\label{asia}
We consider a small synthetic problem called Asia \citep{lauritzen1988local} as our target Bayesian network structure learning problem.
The Asia dataset is composed of 5,000 samples, each is generated by a true network with 8 binary variables\footnote{http://www.bnlearn.com/documentation/man/asia.html}. Bayesian Information Criteria (\BIC) score is used to evaluate how well a Bayesian network fits the 5,000 samples. To train a \VAE model to generate Bayesian network structures, we sample 200,000 random 8-node Bayesian networks from the \texttt{bnlearn} package \citep{JSSv035i03} in R, which are split into 90\% training and 10\% testing sets. Our task is to train a \VAE model on the training Bayesian networks, and search in the latent space for Bayesian networks with high \BIC scores using Bayesian optimization. In this task, we consider a simplified case where the topological order of the true network is known -- we let the sampled training and test Bayesian networks have topological orders consistent with the true network of Asia. This is a reasonable assumption for many practical applications, e.g., when the variables have a temporal order \citep{koller2009probabilistic}. When sampling a network, the probability of a node having an edge with a previous node (as specified by the order) is set to the default option $2/(k-1)$, where $k=8$ is the number of nodes, which results in sparse graphs where the number of edges is in the same order of the number of nodes.



%



\section{Baselines}
\label{baselines}
As discussed in the related work, there are other types of graph generative models that can potentially work for \DAGs. We explore three possible approaches and contrast them with \DVAE.

\noindent\textbf{S-VAE.} The \SVAE baseline treats a \DAG as a sequence of node strings, which we call string-based variational autoencoder (\SVAE). In \SVAE, each node is represented as the one-hot encoding of its type number concatenated with a 0/1 indicator vector indicating which previous nodes have directed edges to it (i.e., a column of the adjacency matrix). For example, suppose there are two node types and five nodes, then node 4's string ``0 1, 0 1 1 0 0'' means this node has type 2, and has directed edges from previous nodes 2 and 3. \SVAE leverages a standard \GRU-based \RNN variational autoencoder \cite{bowman2015generating} on the topologically sorted node sequences, with each node's string treated as its input bit vector. 

\noindent\textbf{GraphRNN.} One similar generative model is \GraphRNN \citep{you2018graphrnn}. Different from \SVAE, it further decomposes an adjacency column into entries and generates the entries one by one using another edge-level \GRU. \GraphRNN is a pure generative model which does not have an encoder, thus cannot optimize \DAG performance in a latent space. To compare with \GraphRNN, we equip it with \SVAE's encoder and use it as another baseline. Note that the original \GraphRNN feeds nodes using a BFS order (for undirected graphs), yet we find that it is much worse than using a topological order here. Note also that although \GraphRNN seems more expressive than \SVAE, we find that in our applications \GraphRNN tends to have more severe overfitting and generates less diverse \DAGs.


\begin{wrapfigure}[8]{L}{0.6\textwidth}
\centering
\vspace{-10pt}
\includegraphics[width=0.6\textwidth]{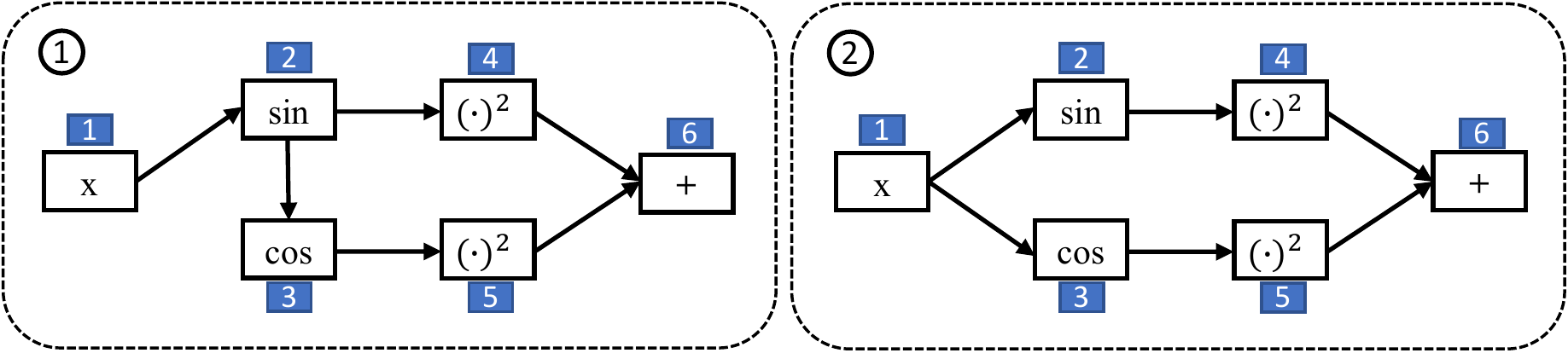}
\caption{\small Two bits of change in the string representations can completely change the computational purpose. 
}
\label{SVAE_brittle}
\end{wrapfigure}

Both \GraphRNN and \SVAE treat \DAGs as bit strings and use \RNNs to model them. This representation has several drawbacks. Firstly, since the topological ordering is often not unique for a \DAG, there might be multiple string representations for the same \DAG, which all result in different encoded representations. This will violate the permutation invariance in Theorem \ref{perm_inv}. Secondly, the string representations can be very brittle in terms of modeling \DAGs' computational purposes. In Figure \ref{SVAE_brittle}, the left and right \DAGs' string representations are only different by two bits, i.e., the edge (2,3) in the left is changed to the edge (1,3) in the right. However, the two bits of change in structure greatly changes the signal flow, which makes the right \DAG always output $1$.
In \SVAE and \GraphRNN, since the bit representations of the left and right \DAGs are very similar, they are highly likely to be encoded to similar latent vectors. In particular, the only difference between encoding the left and right \DAGs is that, for node 3, the encoder \RNN will read an adjacency column of [0, 1, 0, 0, 0, 0] in the left, and read [1, 0, 0, 0, 0, 0] in the right, while all the remaining encoding is exactly the same. 
By embedding two \DAGs serving very different computational purposes to the same region of the latent space, \SVAE and \GraphRNN tend to have less smooth latent spaces which make optimization on them more difficult. In contrast, \DVAE can better differentiate such subtle differences, as the change of edge (2,3) to (1,3) completely changes what aggregated message node 3 receives in \DVAE (hidden state of node 2 vs. hidden state of node 1), which greatly affects node 3 and all its successors' feature learning.

\noindent\textbf{GCN.} The graph convolutional network (\GCN) \cite{kipf2016semi} is one representative graph neural network with a simultaneous message passing scheme. In \GCN, all the nodes take their neighbors' incoming messages to update their own states simultaneously instead of following an order. After message passing, the summed node states is used as the graph state. We include \GCN as the third baseline. Since \GCN can only encode graphs, we equip \GCN with \DVAE's decoder to make it a \VAE model. For neural architectures, we searched the number of message passing layers from 1 to 5. We found that if we only use 1 message passing layer, the reconstruction accuracy is only around 5\%. And if we use 2 or more layers, the reconstruction accuracy gets around 97\% stably but never reaches nearly 100\% like other models. This demonstrates \GCN's limitation of only encoding local substructures for neural architectures. The final \GCN model uses 3 message passing layers. For Bayesian networks, we find 1 layer is enough to reach a 99\% reconstruction accuracy, which is reasonable since Bayesian networks are naturally local. We report a \GCN model using 2 message passing layers.

Using \GCN as the encoder can ensure permutation invariance, since node ordering does not matter in \GCN. However, \GCN's message passing focuses on propagating the neighboring nodes' features to each center node to encode the \textbf{local substructure pattern} around each node. In comparison, \DVAE's message passing simulates how the computation is performed along the directed paths of a \DAG and focuses on encoding the computation. Although learning local structural features is essential for \GCN's successes in node classification and graph classification, here in our tasks, modeling the computation represented by the entire graph is much more important than modeling the local features. Encoding only local substructures may also lose important information about the global \DAG topology, making it more difficult to reconstruct the \DAG.

\noindent\textbf{DeepGMG.} \dgmg \citep{li2018learning} is a graph-based graph generative model that uses a simultaneous message passing to learn intermediate node/graph states and uses a similar decoding scheme to \DVAE to generate nodes/edges of a graph sequentially. \dgmg is originally designed for generating general (undirected) graphs.
Several modifications are made to adapt it to our tasks. 
First, we make it a \VAE by equipping it with a 3-layer message passing network as the encoder using its own message passing functions, and use the original generative model as the decoder. 
Second, we feed in nodes using a topo-order instead of the original random order (and see much improvement). 
Third, we make the sampled edges in the decoding phase only point to new nodes to ensure acyclicity.

Similar to \GCN, \dgmg's training loss never reaches near zero even with extensive hyperparameter tuning, which again reveals the limitation of simultatenous message passing for encoding \DAGs. In comparison, \DVAE can be perfectly trained to near zero loss.  

We omit other possible approaches such as \acro{G}raph\acro{VAE} \cite{simonovsky2018graphvae} and some recent graph-based models \cite{jin2018junction,liu2018constrained,you2018graph} etc., either because they lack official code or they target specific graphs (such as molecules) only.

\section{VAE Training Details}\label{appendix:training}

We use the same settings and hyperparameters (where applicable) for all the four models to be as pair as possible. Many hyperparameters are inherited from \citet{kusner2017grammar}. Single-layer \GRUs are used in all models requiring recurrent units, with the same hidden state size of 501. We set the dimension of the latent space to be 56 for all models. All \VAE models use $\mathcal{N}(\mathbf{0}, \mat{I})$ as the prior distribution $p(\vec{z})$, and take $q_\phi(\vec{z}|G)$ ($G$ denotes the input \DAG) to be a normal distribution with a diagonal covariance matrix, whose mean and variance parameters are output by the encoder. The two \MLPs used to output the mean and variance parameters are all implemented as single linear layer networks.

For the decoder network of \DVAE, we let $f_{\text{add\_vertex}}$ and $f_{\text{add\_edge}}$ be two-layer \MLPs with ReLU nonlinearities, where the hidden layer sizes are set to two times of the input sizes. Softmax activation is used after $f_{\text{add\_vertex}}$, and sigmoid activation is used after $f_{\text{add\_edge}}$. For the gating network $g$, we use a single linear layer with sigmoid activation. For the mapping function $m$, we use a linear mapping without activation. The bidirectional encoding discussed in Section \ref{discussion} is enabled for \DVAE on neural architectures, and disabled for \DVAE on Bayesian networks and other models where it gets no better results. 

When optimizing the \VAE loss, we use $\textrm{ReconstructLoss}+\alpha\textrm{KLDivergence}$ as the loss function. In the original \VAE framework, $\alpha$ is set to 1. However, we found that it led to poor reconstruction accuracies, similar to the findings of previous work \cite{kusner2017grammar,dai2018syntax,jin2018junction}. Following the implementation of \citet{jin2018junction}, we set $\alpha=0.005$. Mini-batch SGD with Adam optimizer \cite{kingma2014adam} is used for all models. For neural architectures, we use a batch size of 32 and train all models except \dgmg for 300 epochs. For Bayesian networks, we use a batch size of 128 and train all models except \dgmg for 100 epochs. For \dgmg, we early stop the training at epoch 30 and epoch 5 for neural architectures and Bayesian networks, respectively, in order to avoid numerical instabilities. We use an initial learning rate of 1E-4, and multiply the learning rate by 0.1 whenever the training loss does not decrease for 10 epochs. We use PyTorch to implement all the models.

\section{SGP Training Details}\label{sgp}
We use sparse Gaussian process (\SGP) regression as the predictive model. We use the open sourced \SGP implementation in \cite{kusner2017grammar}. Both the training and testing data's performances are standardized according to the mean and std of the training data's performances before feeding to the \SGP. And the \RMSE and Pearson's $r$ in Table \ref{results2} are also calculated on the standardized performances. We use the default Adam optimizer to train the \SGP for 100 epochs constantly with a mini-batch size of 1,000 and learning rate of 5E-4.

For neural architectures, we use all the training data to train the \SGP. For Bayesian networks, we randomly sample 5,000 training examples each time, due to two reasons: 1) using all the 180,000 examples to train the \SGP might not be realistic for a typical scenario where network/dataset is large and evaluating a network is expensive; and 2) we found using a smaller sample of training data results in more stable \bo performance due to the less probability of duplicate rows which might result in ill conditioned matrices. Note also that, when training the variational autoencoders, all the training data are used, since the \VAE training is purely unsupervised.

\section{More Experimental Results}\label{appendix:experiment}

\subsection{More details on the piror validity experiment}\label{validity}
Since different models can have different levels of convergence w.r.t. the KLD loss in (\ref{VAE_loss}), their posterior distribution $q_\phi(\vec{z}\given \vec{x})$ may have different degrees of alignment with the prior distribution $p(\vec{z}) = \mathcal{N}(\mathbf{0}, \mat{I})$. If we evaluate prior validity by sampling from $p(\vec{z})$ for all models, we will favor those models that have a higher-level of KLD convergence. To remove such effects and focus purely on models' intrinsic ability to generate valid \DAGs, when evaluating prior validity, we apply $\vec{z} = \vec{z} \odot \text{std}(\mat{Z}_{\text{train}}) + \text{mean}(\mat{Z}_{\text{train}})$ for each model (where $\mat{Z}_{\text{train}}$ are encoded means of the training data by the model), so that the latent vectors are scaled and shifted to the center of the training data's embeddings. If we do not apply such transformations, we find that we can easily control the prior validity results by optimizing for more or less epochs or putting more or less weight on the KLD loss.

For a generated neural architecture to be read by \ENAS, it has to pass the following validity checks: 1) It has one and only one starting node (the input layer); 2) It has one and only one ending type (the output layer); 3) Other than the input node, there are no nodes which do not have any predecessors (no isolated paths); 4) Other than the output node, there are no nodes which do not have any successors (no blocked paths); 5) Each node must have a directed edge from the node immediately before it (the constraint of \ENAS), i.e., there is always a main path connecting all the nodes; and 6) It is a \DAG.

For a generated Bayesian network to be read by \texttt{bnlearn} and evaluated on the Asia dataset, it has to pass the following validity checks: 1) It has exactly 8 nodes; 2) Each type in "ASTLBEXD" appears exactly once; and 3) It is a \DAG.

Note that the training graphs generated by the original software all satisfy these validity constraints.

\begin{figure*}[!htb]
\centering
\includegraphics[width=0.45\textwidth]{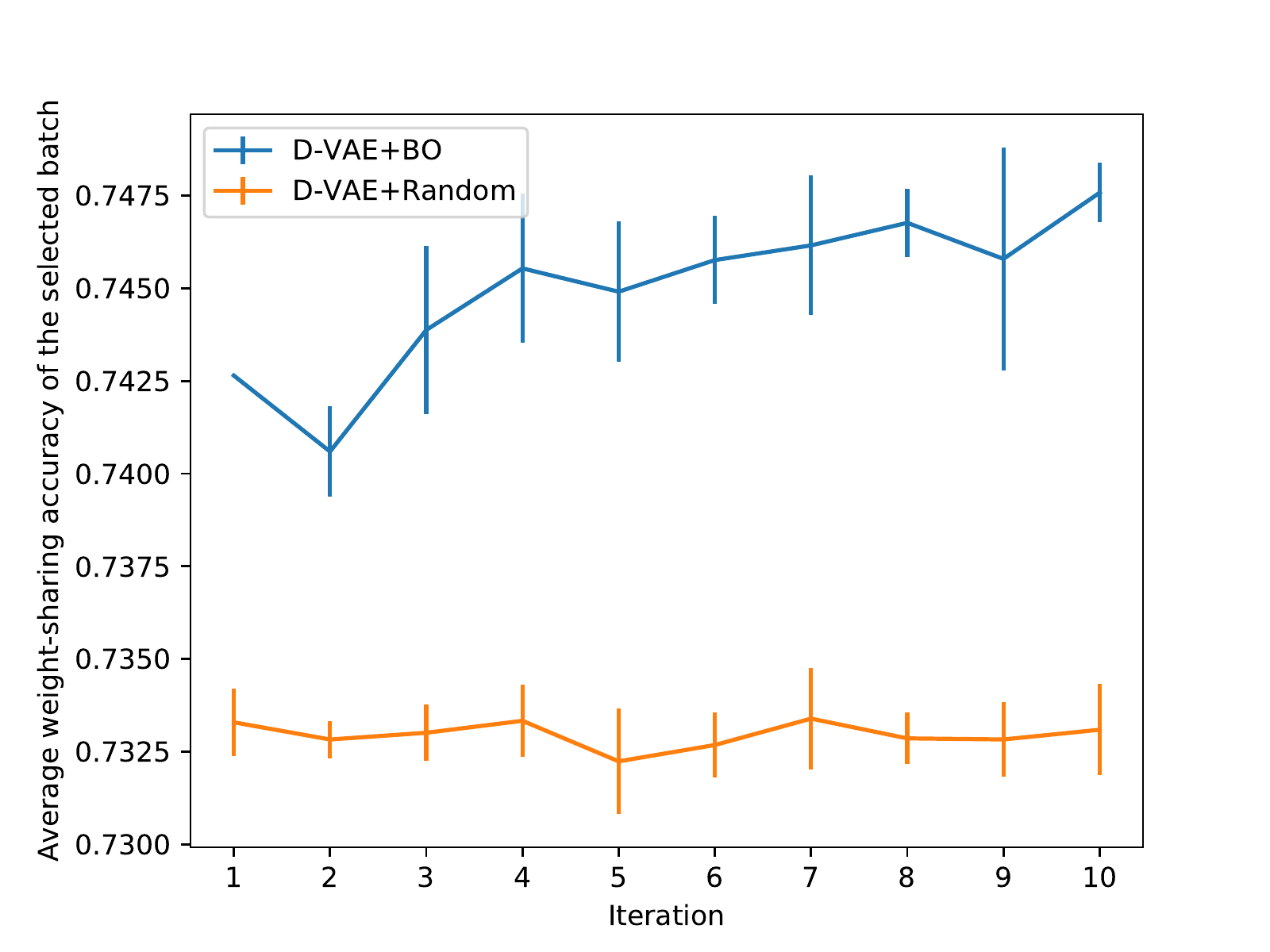}
\includegraphics[width=0.45\textwidth]{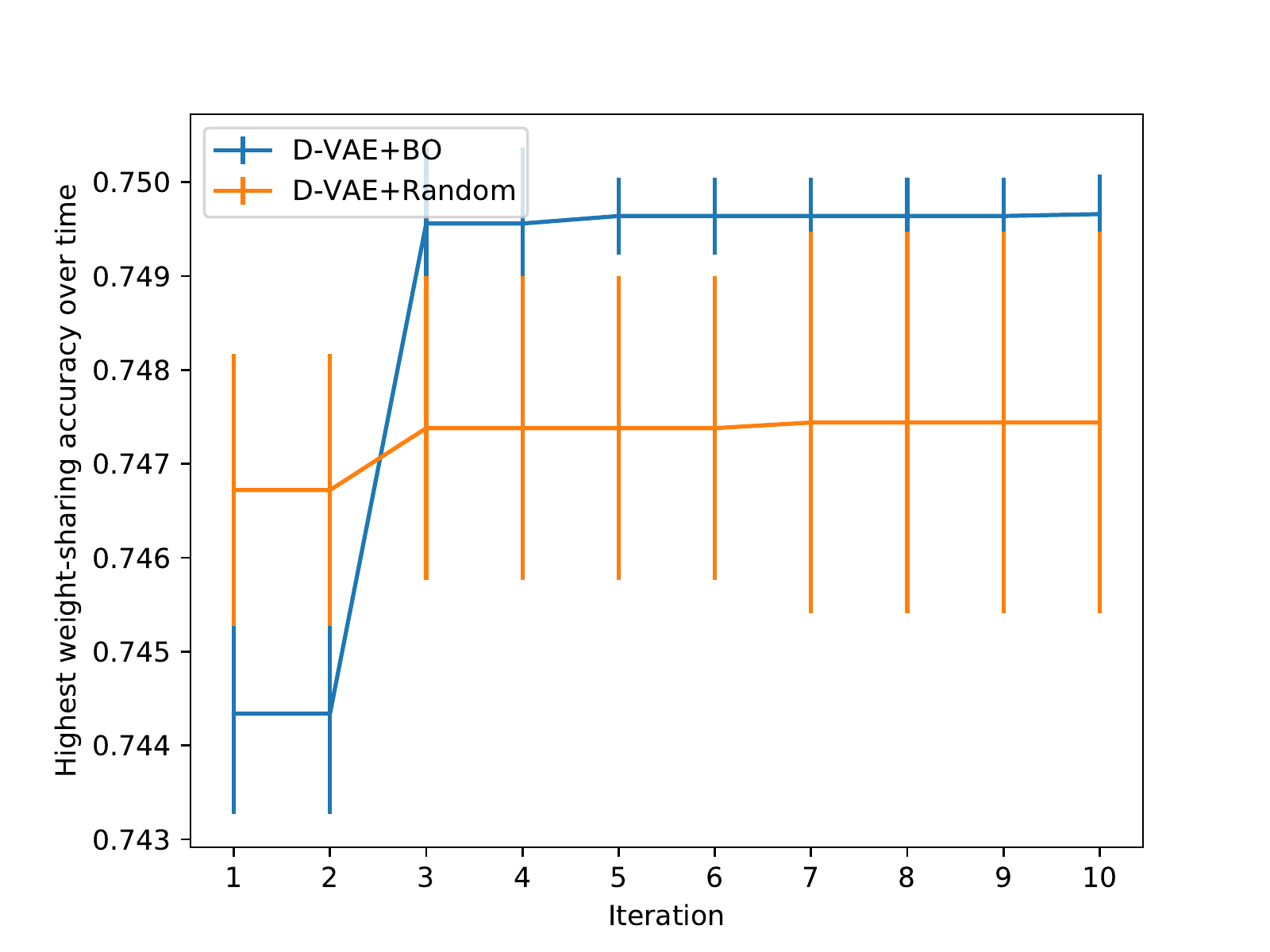}
\caption{\small Comparing \bo with random search on neural architectures. Left: average weight-sharing accuracy of the selected points in each iteration. Right: highest weight-sharing accuracy of the selected points over time.}
\label{bo_random}
\includegraphics[width=0.45\textwidth]{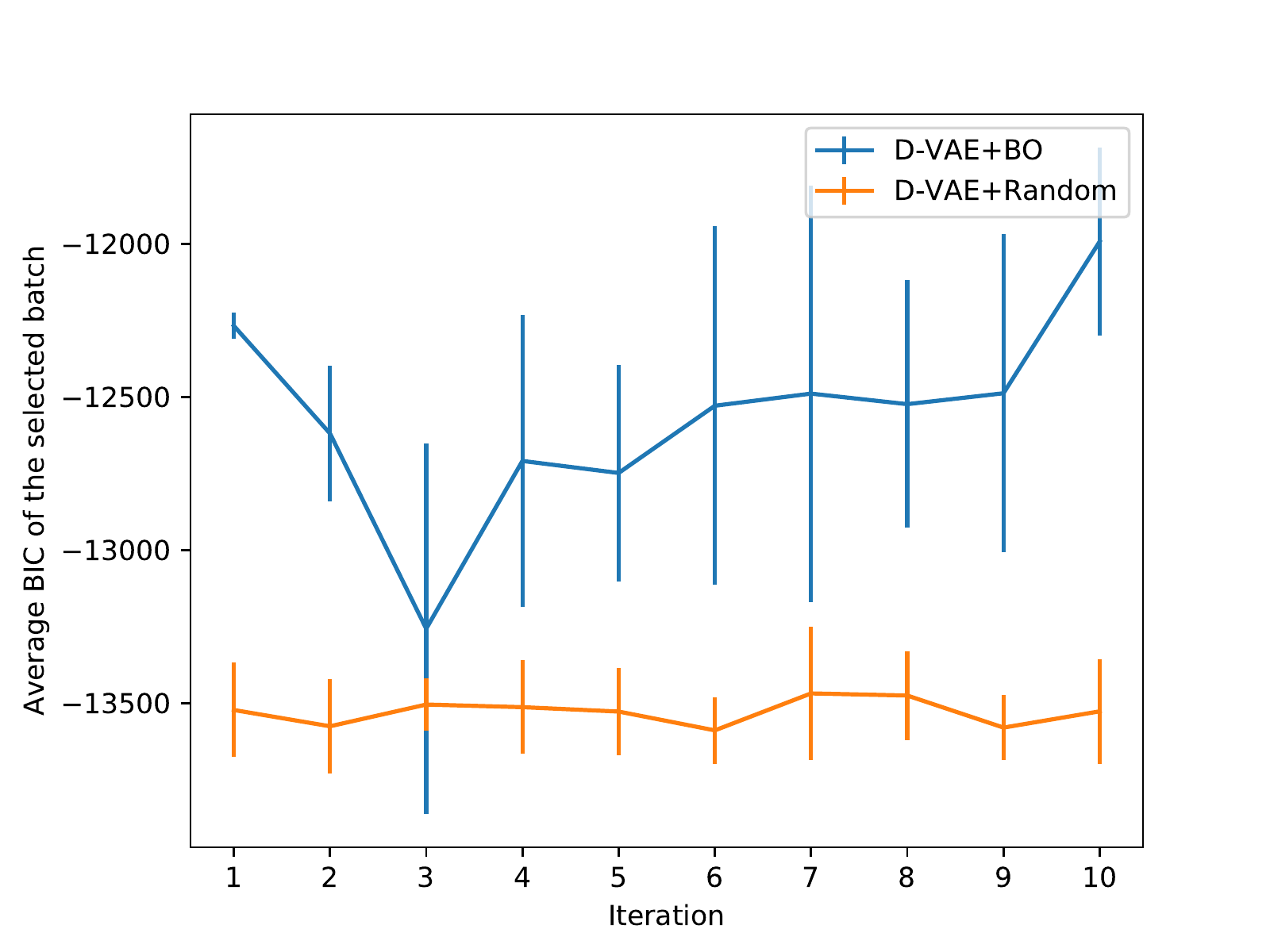}
\includegraphics[width=0.45\textwidth]{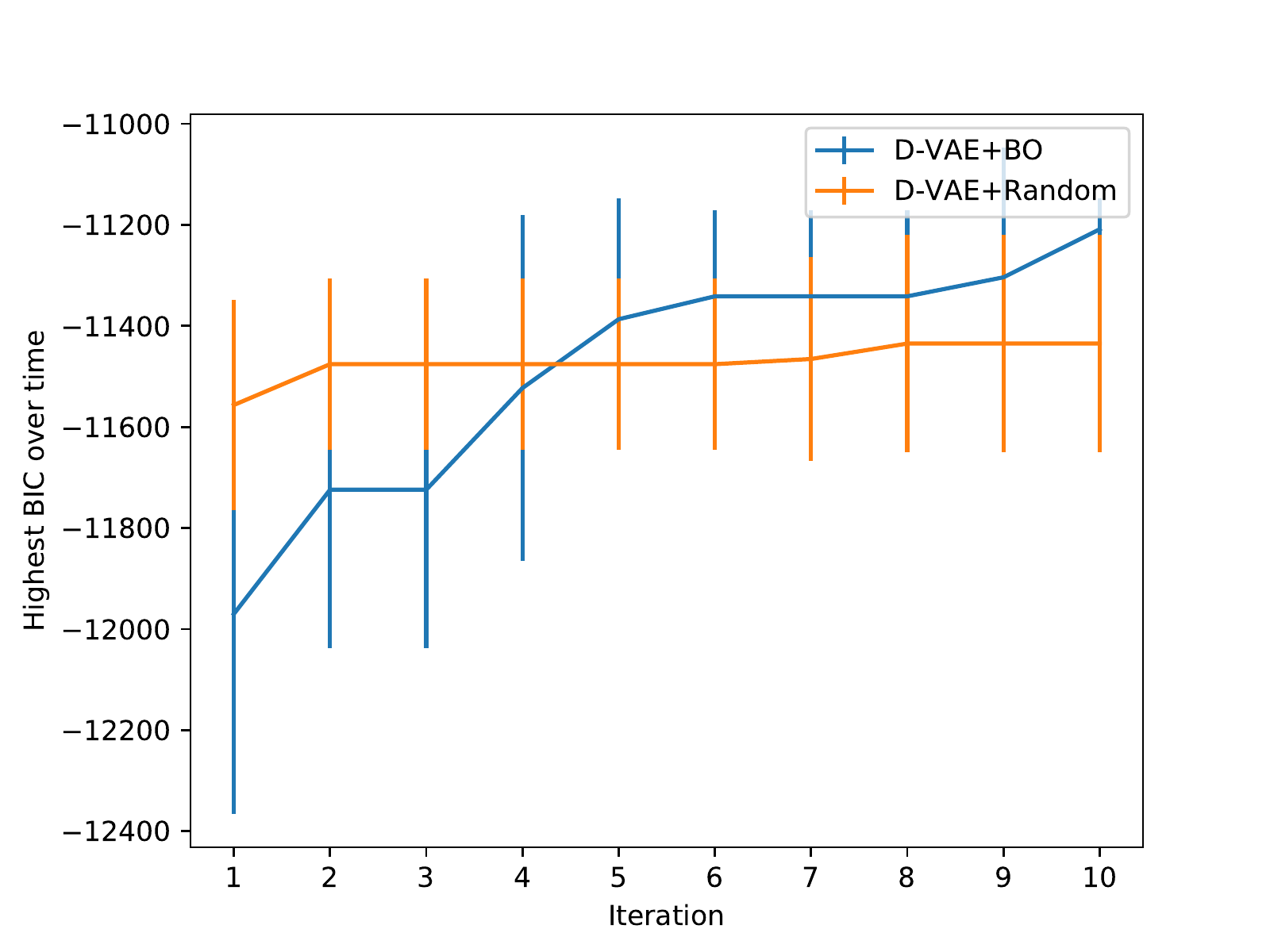}
\caption{\small Comparing \bo with random search on Bayesian networks. Left: average \BIC score of the selected points in each iteration. Right: highest \BIC score of the selected points over time.}
\label{bo_random_BN}
\end{figure*}

\subsection{Bayesian optimization vs. random search}
To validate that Bayesian optimization (\bo) in the latent space does provide guidance in searching better \DAGs, we compare \bo with Random (which randomly samples points from the latent space of \DVAE). Figure \ref{bo_random} and \ref{bo_random_BN} show the results (averaged across 10 trials). In each figure, the left plot shows the average performance of all the points found in each \bo round, and the right plot shows the highest performance of all the points found so far.
As we can see, \bo consistently selects points with better average performance in each round than random search, which is expected. However, for the highest performance results, \bo tends to fall behind Random in the initial few rounds. This might be because our batch expected improvement heuristic aims to take advantage of the currently most promising regions by selecting most points of the batch in the same region (exploitation), while Random more evenly explores the entire space (exploration). Nevertheless, \bo seems to quickly catch up after a few rounds and shows long-term advantages.

\subsection{More D-VAE experiments}
\DVAE leverages the proposed asynchronous message passing in both its encoder and decoder. To understand deeper how the asynchronous message passing helps, we add some ablation studies of \DVAE on our neural architecture datasets. Firstly, we replace the asynchronous message passing in \DVAE with simultaneous message passing to construct a \DVAE \acro{(SMP)} baseline. Secondly, we keep the \DVAE encoder unchanged, and replace the decoder with \SVAE's string-based decoder. The decoder now is a simple \RNN with $\mathcal{O}(n)$ complexity instead of the original $\mathcal{O}(n^2)$ complexity, thus is much faster. We name this variant \DVAE \acro{(fast)}. We compare these two variants with the original \DVAE and \SVAE on our 6-layer neural architectures. The results are shown in Table \ref{ablation1}. We can see that \DVAE still in general has the best performance. The \DVAE \acro{(SMP)} baseline shows inferior reconstruction accuracy due to its nonzero training loss caused by the simultaneous message passing. The \DVAE \acro{(fast)} shows similar generative ability to \DVAE. In terms of latent space predictive ability, it is inferior to \DVAE but better than \SVAE. This indicates that, it is beneficial to use asynchronous message passing in both \DVAE's encoder and decoder, rather than only using it to encode \DAGs.

\begin{table}[t]
\centering
\caption{\small Generative ability and latent space predictive ability of \DVAE and its variants.}
\resizebox{0.86\textwidth}{!}{
\begin{tabular}{@{}lcccccccc@{}}
\toprule
        & \multicolumn{4}{c}{Generative ability (\%)}    & \multicolumn{2}{c}{Predictive ability}       \\ \cmidrule(r{0.5em}){2-5} \cmidrule(l{0.5em}){6-7}
Methods & Accuracy   & Validity    & Uniqueness  & Novelty     &  RMSE         & Pearson's $r$  \\ \midrule
D-VAE (SMP) & 92.35    & 99.75   & \textbf{65.98}   &  \textbf{100.00} & 0.455$\pm$0.002 &  0.885$\pm$0.001  \\
D-VAE (FAST)  & \textbf{99.98}   & \textbf{100.00}   & 40.53   & \textbf{100.00}  & 0.419$\pm$0.006 & 0.905$\pm$0.001     \\
D-VAE      & \textbf{99.96}   & \textbf{100.00}   & 37.26   & \textbf{100.00}  & \textbf{0.384$\pm$0.002} & \textbf{0.920$\pm$0.001}     \\
S-VAE    & \textbf{99.98}   & \textbf{100.00}   & 37.03   & \textbf{99.99}  & 0.478$\pm$0.002     & 0.873$\pm$0.001     \\ 
\bottomrule
\end{tabular}
}
\label{ablation1}
\end{table}

\begin{figure*}[h]
\centering
\includegraphics[width=0.25\textwidth]{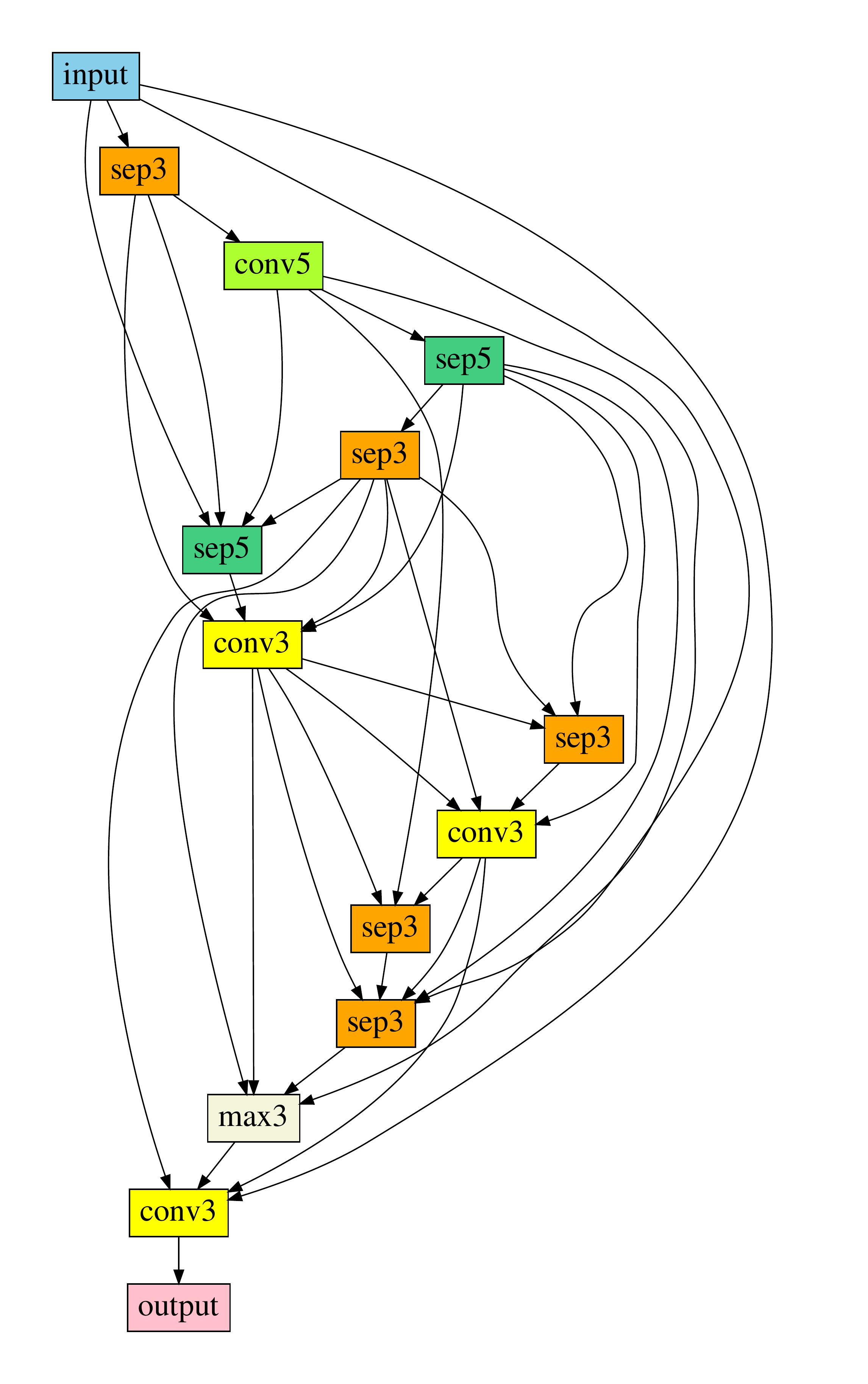}
\caption{Visualization of the final 12-layer neural architecture found by \DVAE \acro{(fast)}.}
\label{nn12}
\end{figure*}

Nevertheless, using \DVAE \acro{(fast)} allows us to work on deeper neural architectures with much less training time due to its linear decoding complexity. Thus, we repeat our \NAS experiments on 12-layer neural architectures. Our final found network has an error rate of 3.88\%, comparable to many state-of-the-art \NAS results in the macro space such as \citep{pham2018efficient}. We plot our final 12-layer neural architecture in Figure \ref{nn12}.

\subsection{More visualization results for neural architectures}

\begin{figure*}[!t]
\centering
\fbox{\includegraphics[width=0.47\textwidth]{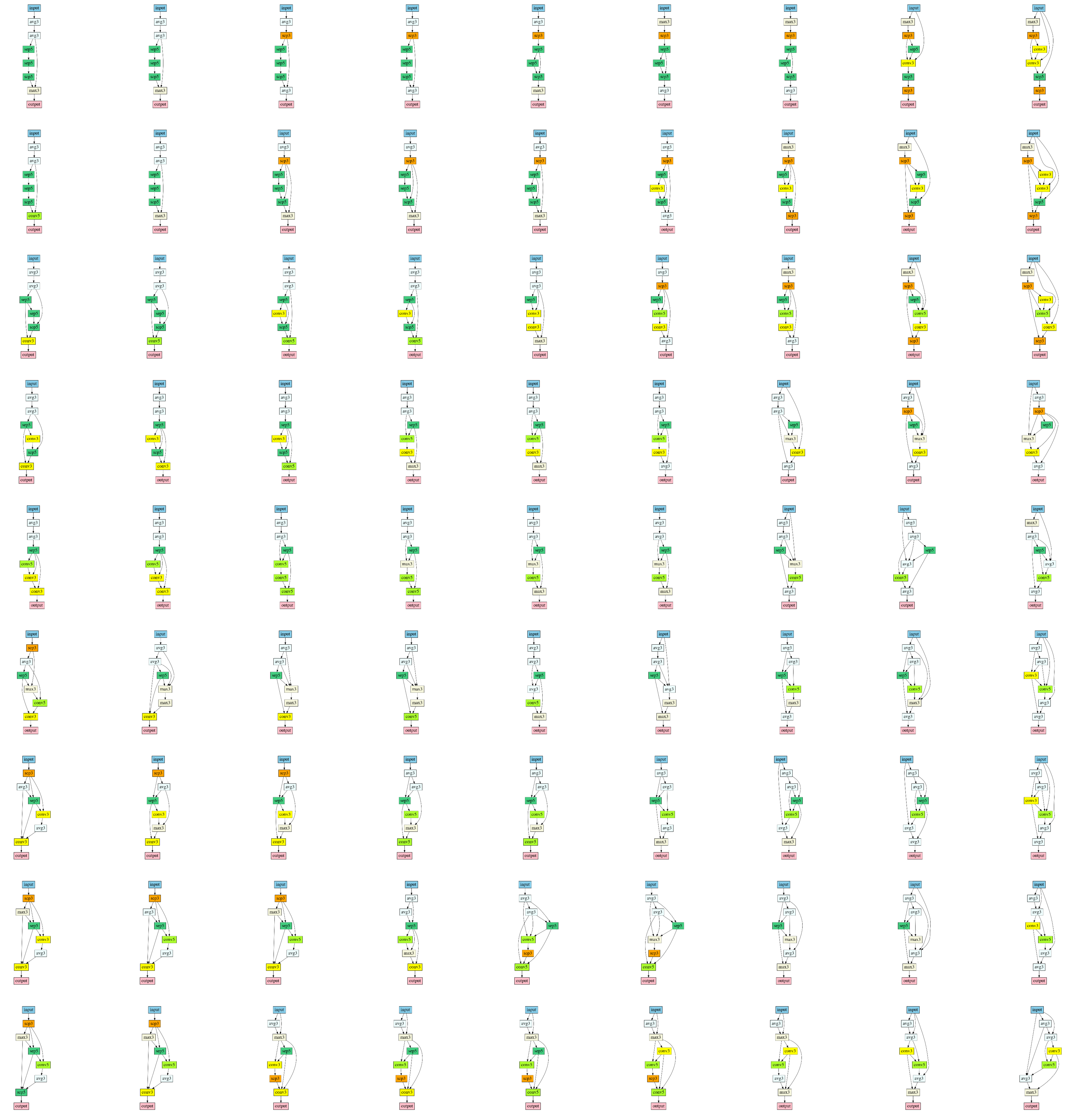}}
\hspace{\fill}
\fbox{\includegraphics[width=0.475\textwidth]{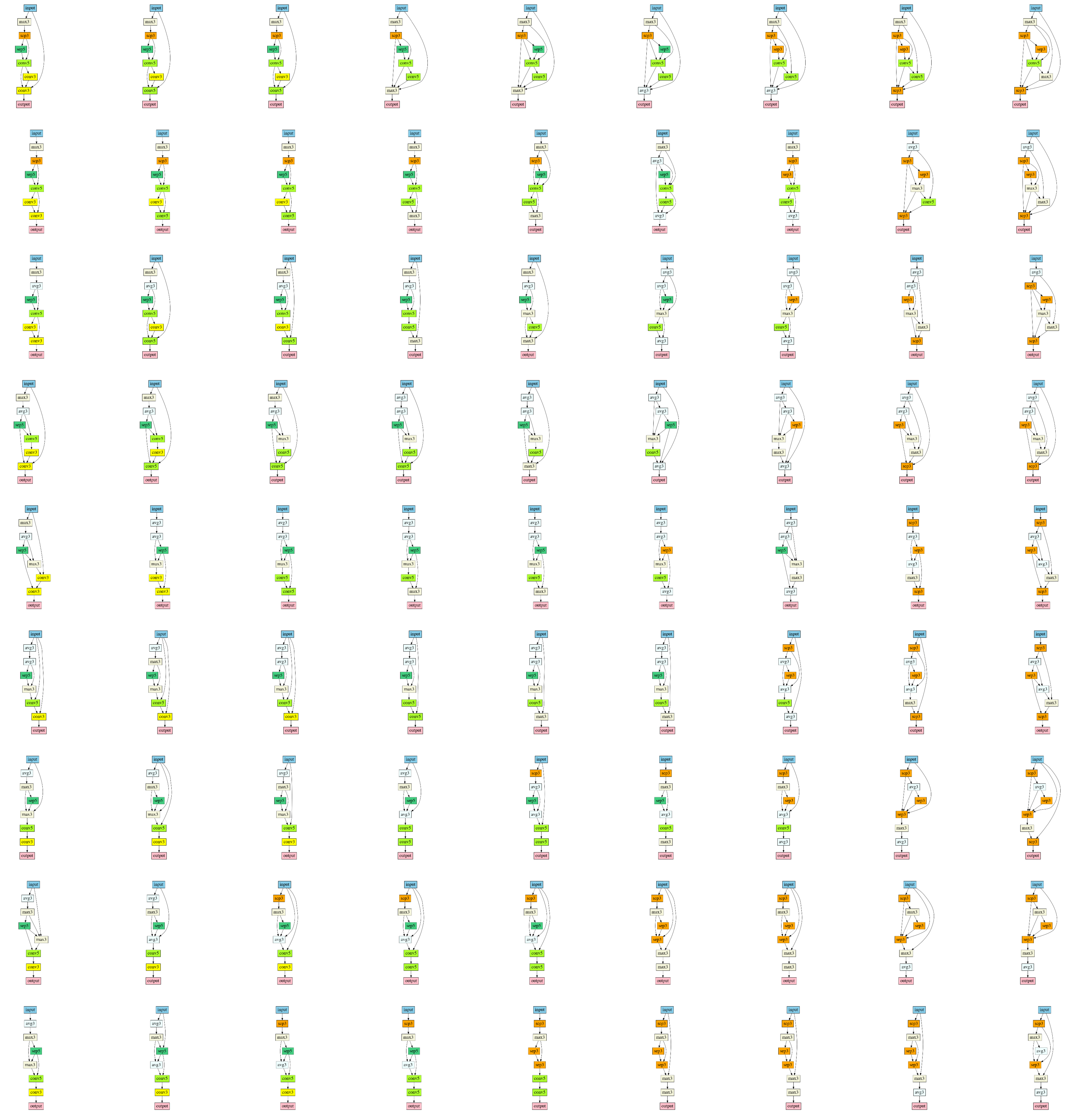}}
\caption{\small 2-D visualization of decoded neural architectures. Left: \DVAE. Right: \SVAE.}
\label{NN_smoothness}
\end{figure*}

We randomly pick a neural architecture and use its encoded mean as the starting point. We then generate two random orthogonal directions, and move in the combination of these two directions from the starting point to render a 2-D visualization of the decoded architectures in Figure \ref{NN_smoothness}.

\begin{figure*}[h]
\centering
\fbox{\includegraphics[width=0.475\textwidth]{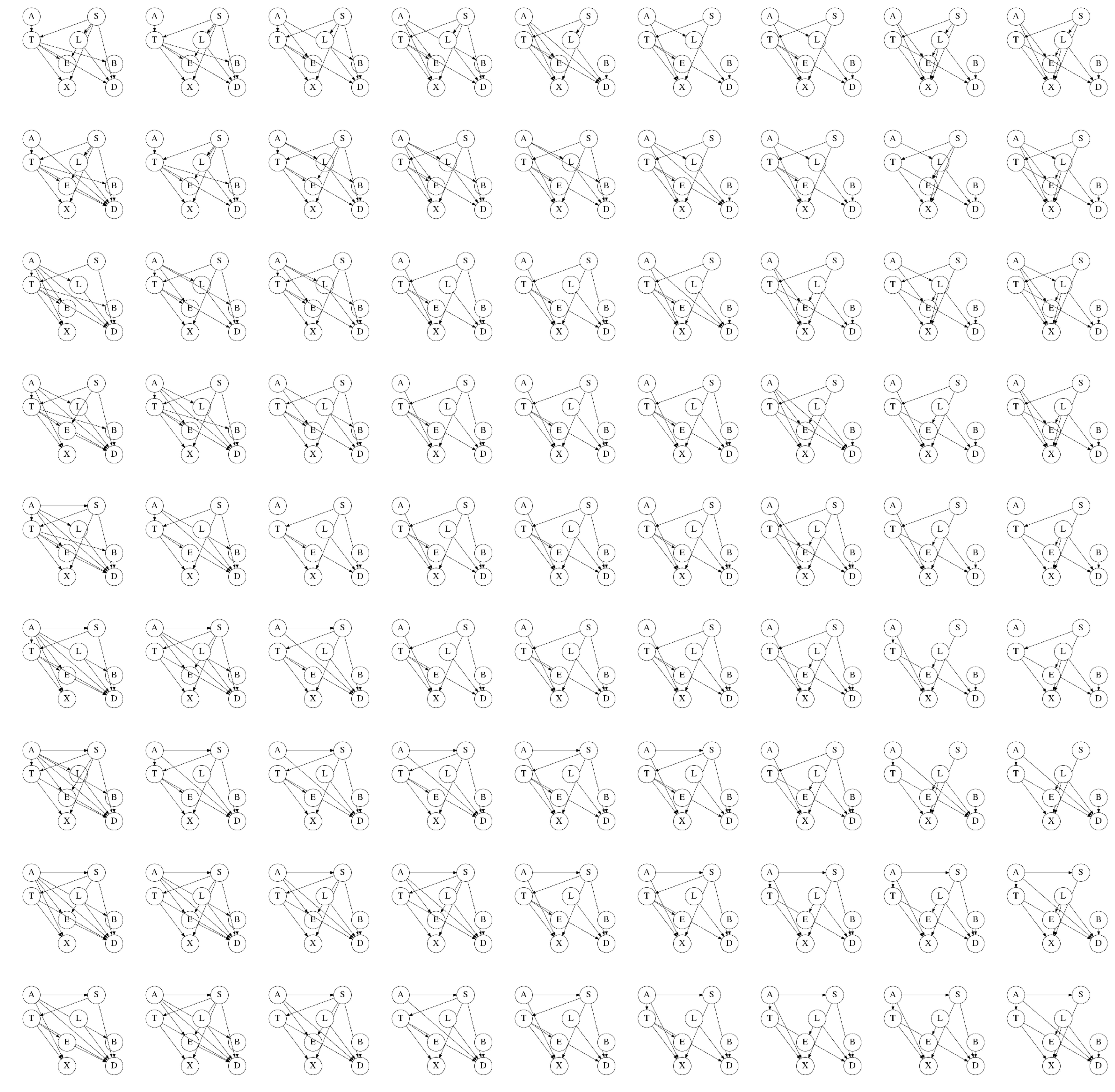}}
\hspace{\fill}
\fbox{\includegraphics[width=0.475\textwidth]{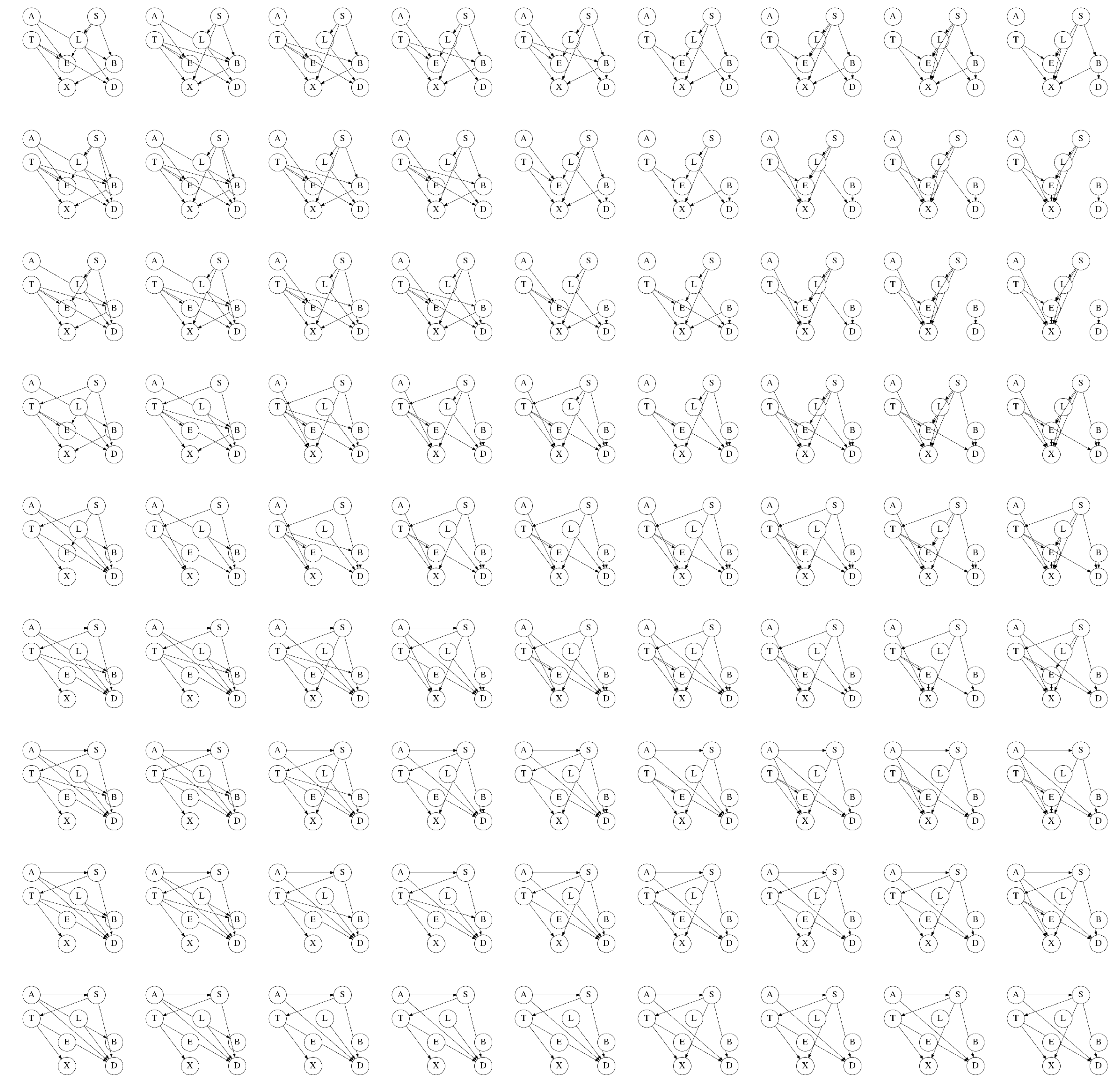}}
\caption{\small 2-D visualization of decoded Bayesian networks. Left: \DVAE. Right: \SVAE.}
\label{BN_smoothness}
\end{figure*}

\subsection{More visualization results for Bayesian networks}
We similarly show the 2-D visualization of decoded Bayesian networks in Figure \ref{BN_smoothness}. Both \DVAE and \SVAE show smooth latent spaces.



\end{appendices}

\end{document}